\documentclass[twoside]{article}

 \usepackage[accepted]{aistats2024}

\usepackage{hyperref}
\usepackage{url}

\usepackage[utf8]{inputenc} 
\usepackage[T1]{fontenc}    
\usepackage{hyperref}       
\usepackage{url}            
\usepackage{booktabs}       
\usepackage{amsfonts}       
\usepackage{multirow, makecell}
\usepackage{nicefrac}       
\usepackage{microtype}      
\usepackage{graphicx}
\usepackage{xcolor, colortbl}
\usepackage{soul}
\usepackage{wrapfig}
\definecolor{LightCyan}{rgb}{0.88,1,1}
\newcolumntype{g}{>{\columncolor{gray!25}}c}

\usepackage{algorithm} 
\usepackage{algpseudocode}
\usepackage{lmodern}
\usepackage{amsthm}
\usepackage{amsmath}
\usepackage{mathtools}
\usepackage{setspace}
\usepackage{caption} 
\usepackage{subcaption}
\usepackage{esvect}
\usepackage{amssymb}
\usepackage[textsize=tiny,color=gray!10]{todonotes}

\newcommand\ours{{Linear Calibration}}
\newlength{\mysize}
\newcommand{\mycfs}[1]{\setlength{\mysize}{#1pt}%
  \fontsize{\mysize}{1.2\mysize}\selectfont}

\begin{document}

%

%

\twocolumn[

\aistatstitle{Enhancing In-context Learning via Linear Probe Calibration}

\vspace{-0.5cm}
\aistatsauthor{Momin Abbas$^{\star}$ \And Yi Zhou$^{\dagger}$ \And Parikshit Ram$^{\dagger}$} 
\vspace{0.1cm}
\aistatsauthor{Nathalie Baracaldo$^{\dagger}$ \And Horst Samulowitz$^{\dagger}$ \And Theodoros Salonidis$^{\dagger}$ \And Tianyi Chen$^{\star}$ }
\vspace{0.1cm}
\aistatsaddress{$^{\star}$Rensselaer Polytechnic Institute \And  $^{\dagger}$IBM  Research  } 
]

\begin{abstract}
In-context learning (ICL) is a new paradigm for natural language processing that utilizes Generative Pre-trained Transformer (GPT)-like models. This approach uses prompts that include in-context demonstrations to generate the corresponding output for a new query input. However, applying ICL in real cases does not scale with the number of samples, and lacks robustness to different prompt templates and demonstration permutations. In this paper, we first show that GPT-like models using ICL result in 
unreliable predictions based on a new metric based on Shannon entropy. 
Then, to solve this problem, we propose a new technique called the \textit{Linear Probe Calibration} (LinC), a method that calibrates the model's output probabilities, resulting in reliable predictions and improved performance, while requiring only minimal additional samples (as few as five labeled data samples). LinC significantly enhances the ICL test performance of GPT models on various benchmark datasets, with an average improvement of up to 21\%, and up to a 50\% improvement in some cases, and significantly boosts the performance of PEFT methods, especially in the low resource regime. 
Moreover, LinC achieves lower expected calibration error, and is highly robust to varying label proportions, prompt templates, and demonstration permutations.
 Our code is available at \url{https://github.com/mominabbass/LinC}.
\end{abstract}
\vspace{-0.5cm}
\section{Introduction}
Large language models (LLMs), have remarkably showcased their capabilities across a broad range of natural language processing tasks \cite{devlin2018bert, dong2019unified, GPT3, lewis2019bart, yang2019xlnet, radford2019languagegpt2}.
The cost of training these large models can be prohibitively expensive. Therefore, the commonly adopted approach is to first pre-train with large amounts of unlabled data and then fine-tune the model to downstream tasks. Although fine-tuning LLMs can be effective, it is prone to instability \cite{mosbach2020stability} due to numerous hyperparameter configurations resulting in failed runs, unstable results, and over-fitting \cite{raffel2020exploring, devlin2018bert, kumar2022fine}. {Moreover, fine-tuning models of such large size may also be expensive and also requires explicit access to the architecture and weights of LLMs, which may not be publicly available \cite{zhang2022opt}.}

\begin{figure}[h]
\begin{center}
    \includegraphics[width=0.5\textwidth]{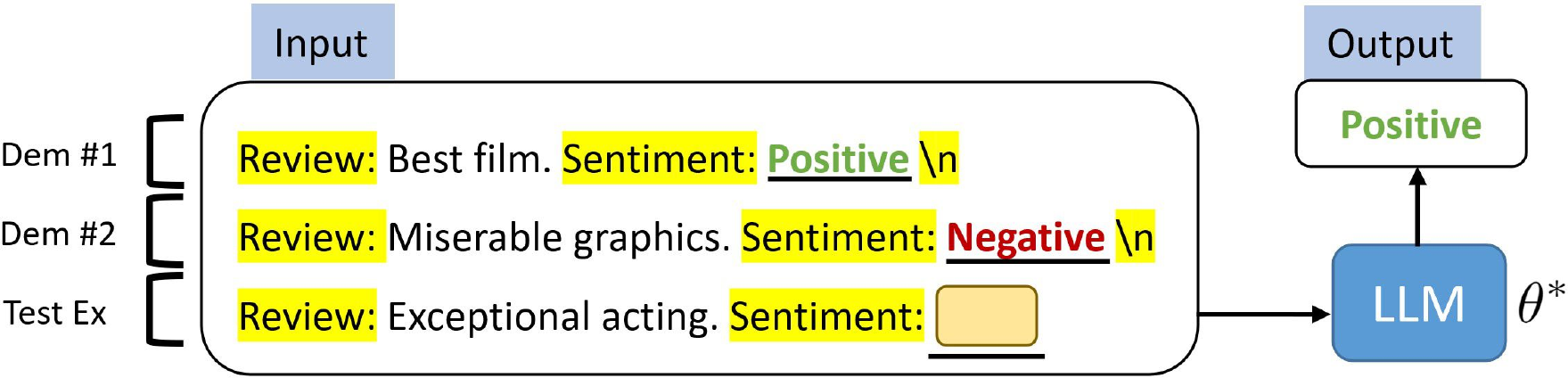}
\end{center}
\vspace{-0.1cm}
\caption{Example of ICL with a LLM $\theta^*$.}
\label{ICL_fig}
\vspace{-0.1cm}
\end{figure}
To avoid large fine-tuning times and avoid requiring access to the weights of the model, recently, LLMs, exemplified by GPT-3 \cite{GPT3}, have demonstrated the ability to perform \emph{in-context learning} (ICL), a capability whereby a model can generate an appropriate output for a given query input based on a prompt that includes input-output example pairs specific to the task at hand. {ICL can work with an API without explicit access to the LLM.}  Figure \ref{ICL_fig} provides a visual representation of ICL. The prompts utilized in ICL entail task-specific in conjunction with a series of input-label pairs, referred to as demonstrations. Such a capability of LLM to learn ``in-context'' presents an intriguing aspect whereby the model is capable of acquiring knowledge and performs well on a wide range of downstream tasks without any task-specific parameter fine-tuning \cite{GPT3, touvron2023llama, black2022gpt, rae2021scaling}.

More specifically, the aim of ICL is to make a prediction on some query test sample $\textbf{x}$ by conditioning on a prompt sequence $(f_{x}(x_1), f_{y}(y_1), \dots, f_{x}(x_k), f_{y}(y_k), f_x(\textbf{x}))$ containing $k$-shot samples $(x_i, y_i)_{i=1}^{k}$ (i.e. demonstrations) and the query test sample, where the functions $f_{x}(.), f_{y}(.)$ denote template functions that attach pre-defined descriptions to the input and output, respectively (c.f. text highlighted in \hl{yellow} in Figure \ref{ICL_fig}). The output template function $f_{y}(.)$, in addition, may transform labels $y_i$ into a natural language format instead into numeric/one-hot labels (e.g. for binary classification transforming labels $(0, 1)$ to $(\sf \textcolor{teal}{Positive}, \textcolor{purple}{Negative})$ (c.f. labels in Figure \ref{ICL_fig})). Mathematically, for an input $\textbf{x}$, a prompt $P$ is defined as:
\begin{align} 
     P(\textbf{x}, (x_i, y_i)_{i=1}^{k})  \triangleq d_{1} \oplus d_{2} \oplus \dots \oplus d_{k} \oplus f_x(\textbf{x})
\end{align}
where each demonstration $d_i$ is given by $f_{x}(x_i) \oplus f_{y} (y_i)$ and $\oplus$ denotes the concatenation operation.

Recent studies suggest that ICL exhibits high variability in performance across different prompt templates, demonstrations, and their arrangement within the prompt arising from biases that favor outputting certain answers \cite{zhao2021calibrate, lu2021fantastically}, resulting in performance variation from random guess to state-of-the-art. 
Further, when the prompt $P$ includes several more demonstrations, ICL's performance is thwarted by the inherent maximum sequence length limitation of the underlying language model.
Moreover, our initial analysis reveals that while GPT-like models' ICL ability delivers acceptable results, their predictions cannot be considered reliable when assessed using Shannon entropy. We will discuss these limitations and challenges of ICL in detail in Section~\ref{sec:shannon}.

In this paper, we argue that by training very few parameters and subsequently utilizing an affine transformation on the output probabilities to \emph{calibrate} the model, the performance, as well as the reliability of these predictions, can be significantly enhanced. We propose \textit{linear probe calibration} (LinC), which optimizes the calibration parameters (i.e. a low-dimensional matrix and vector) using only a few extra samples and minimal computation. Experimental results reveal that LinC significantly outperforms the baselines. Moreover, LinC produces reliable predictions that exhibit consistency across a range of prompt templates and various permutations of demonstrations.

We summarize our contributions below:
\begin{itemize}
\itemsep-0.1em
  \item[\bf C1)] {We present a novel insight that, while GPT-like models often exhibit acceptable ICL performance, their predictions appear to have very low confidence when measured using the Shannon entropy, highlighting a potential cause for their highly variable performance.}
 
  \item[\bf C2)] {We propose \emph{linear probe calibration} (LinC), a simple and black-box method that enhances model's reliability and performance by linearly calibrating output probabilities without requiring any access to model weights or architecture.}

  \item[\bf C3)] We empirically show that LinC consistently outperforms baselines on various benchmark datasets, with a performance boost of up to 21\%, on average, and up to a 50\% improvement in some cases, compared to the vanilla ICL baseline.
\end{itemize}

\section{ICL: Challenges and Opportunities} \label{sec:shannon}
In this section, we will first highlight two key limitations of ICL with current LLMs, which then serve as the motivation of our subsequent algorithm design.

\subsection{Maximum Sequence length Limitation} \label{section_max_seq_len}
\begin{figure}[t]
\begin{center}
    \includegraphics[width=0.4\textwidth]{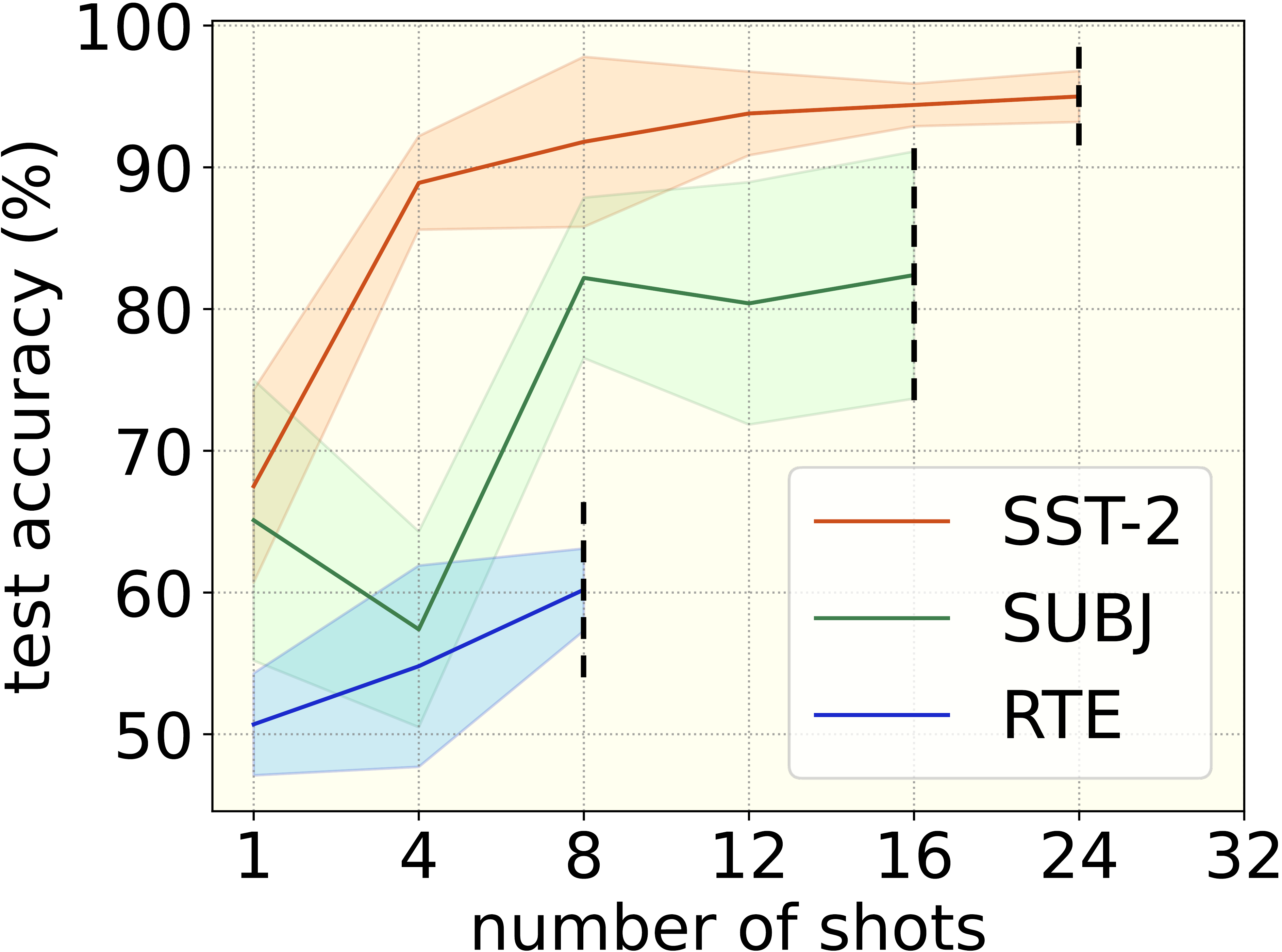}
\end{center}
\vspace{-0.1cm}
\caption{The efficacy of ICL is restricted by GPT tokenizer's maximum sequence length limit. Black-dashed lines demarcate the point beyond which additional shots cannot be utilized.}
  \label{max_seq_len}
  \vspace{-0.5cm}
\end{figure}

We demonstrate the maximum sequence length limitation of ICL \cite{xu2023knn-propmt} on different datasets in Figure \ref{max_seq_len} {using GPT-2 tokenizer}. We observe that the test performance of ICL improves consistently, which aligns with the power-law relations between the generalization error and data size \cite{kaplan2020scaling, rosenfeld2019constructive}. However, beyond a certain point, no additional demonstrations can be added within the prompt since the model reaches its maximum sequence length limit\footnote{It is noteworthy that although some recent models do have expanded context windows (e.g. GPT-4 with 32k tokens), the majority of them still maintain smaller context windows. Smaller models are often preferred in various applications due to their flexibility in adapting to specific tasks and their ability to achieve comparable performance to larger models while using much fewer compute.}.
Alas, the potential for performance enhancement through the acquisition of more examples is often constrained by this limitation of ICL, {even in few-shot settings}. 

Motivated by this, we ask: 
\begin{center}
   \emph{Given {a few} additional samples, {beyond the LLM sequence length limit, can we further improve ICL test performance without utilizing LLM fine-tuning?}}
\end{center}
\begin{figure*}[t]
    \centering
    \includegraphics[width=1.0\textwidth]{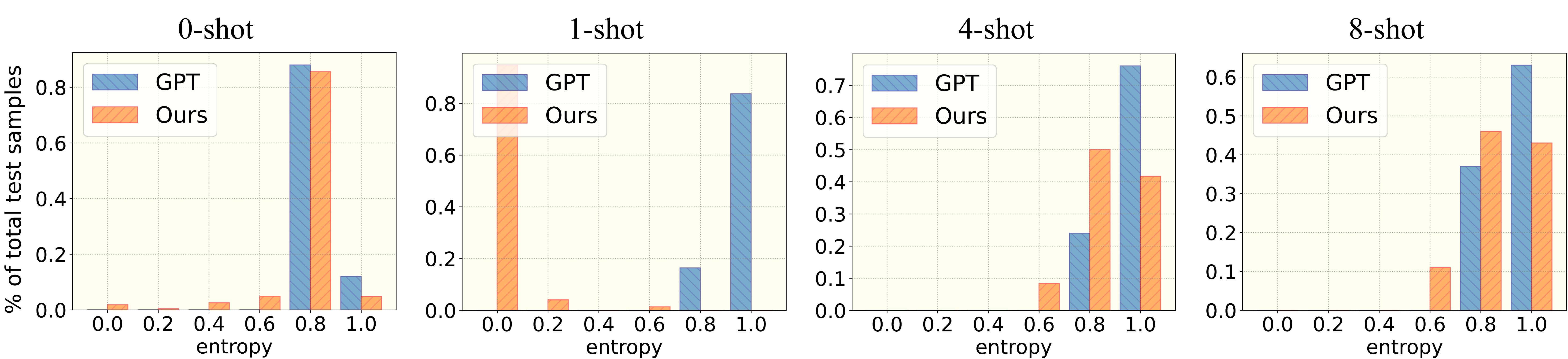}
    \vspace{-0.4cm}
    \caption{Shannon entropy histograms of using vanilla ICL on GPT-2-XL (1.5B) vs our method on SST-2 (higher entropy implies higher uncertainty); we use logarithmic base two. Refer to Section~\ref{sec:shannon} for a detailed explanation.}
    \label{entropy_sst2}
    \vspace{-0.1cm}
\end{figure*}
\subsection{Entropy of Few-Shot Learning with GPT} \label{section_shannon_entropy}
In practical decision-making systems, it is crucial for prediction networks to not only exhibit high accuracy but also have the ability to identify the likelihood of incorrect predictions. For example, automated healthcare systems should be designed such that when the confidence level of a disease diagnosis network is low, the control is transferred to human doctors for making the diagnoses \cite{Jiang2012}. Well-calibrated confidence estimates play a crucial role in making machine learning models more interpretable. Since humans possess an inherent cognitive understanding of probabilities \cite{COSMIDES19961}, having reliable confidence estimates can enhance the user's confidence in the model's predictions. This is especially relevant for neural networks, where the rationale behind their classification decisions can be complex and challenging to decipher.
One way to assess the reliability of a model's predictions is to measure the \textit{Shannon entropy}.
Shannon entropy \cite{shannon1948mathematical} quantifies the amount of expected uncertainty in a probability distribution. Mathematically, Shannon entropy is given by:
\begin{equation} 
     E  = -\sum_{c=1}^{C} \mathbf{p}_c \log (\mathbf{p}_c)
\end{equation}
where $C$ is the number of classes and $\mathbf{p}_c$ represents the output probability of class $c$. Entropy value is used to gauge prediction confidence, with a low value indicating high confidence and vice versa. 
 
We compare the performance of ICL on GPT before (vanilla ICL) and after our calibration (Section \ref{sec:linc}) using different numbers of shots on various datasets. Figure \ref{entropy_sst2} encapsulates the results on SST-2 (for other datasets, see Appendix \ref{Appendix_experiments}). 
We observe that, in contrast to the performance of our model, using vanilla ICL on GPT leads to high values of entropy, implying that most test predictions were made with very low confidence, i.e., close to  random guessing. This observation indicates that although vanilla ICL (i.e., uncalibrated) on GPT yields satisfactory results in terms of test accuracy (refer to Table \ref{table_GPTJ}), the confidence associated with these predictions is not entirely reliable. {These findings align with previous studies, including \cite{han2022prototypical}, that have demonstrated the inadequacy of the conventional GPT decision boundary in effectively distinguishing between predictions by using the output with the highest probability as the predicted label. While low entropy (or high confidence) does not always imply a high accuracy, high entropy typically implies a high degree of uncertainty in predictions. This uncertainty may contribute to the increased variability of test performance, such as variations due to varying label proportions, prompt templates, and demonstration permutations (see Section \ref{section_results}).} Motivated by this, we ask: 
\vspace{-0.2cm}
\begin{center}
   \emph{Can we make the predictions made by GPT-like models  more reliable and accurate?} 
\end{center}%
As we will see in the upcoming sections, our proposal to linearly calibrate model's output probabilities via meticulously learned parameters not only improves its test performance but also enhances its reliability.

\section{Linear Probe Calibration}\label{sec:linc}
In order to be considered reliable, a model must furnish a $\emph{calibrated confidence}$ measure along with its predictions, where the probability assigned to the predicted class corresponds to its actual   likelihood \cite{guo2017calibration}. 

{Since vanilla ICL predictions may be unreliable due to the presence of high entropy prediction probabilities, we aim to investigate whether probability calibration can lead to improvements.} In this section, we present linear probe calibration (LinC) that calibrates the model's output probabilities, making it more  reliable.

One method for modifying output probabilities applies an affine transformation \cite{platt1999, guo2017calibration}:
\begin{equation} \label{eq3}
     \Tilde{\mathbf{p}} = {\sf softmax} (\mathbf{A} \mathbf{p}+\mathbf{b})
\end{equation}
where $\mathbf{A}$ and $\mathbf{b}$ are parameters to be applied to the original probabilities $\mathbf{p}$ to get new probabilities $\Tilde{\mathbf{p}}$. For example, for classification tasks $\mathbf{p}$ is the set of probabilities that are associated with each label\footnote{For detailed information about where the transformation is applied, please refer to Appendix \ref{Appendix_experiments}.}:
\begin{equation} 
     \mathbf{p}(P(\textbf{x}, (x_i, y_i)_{i=1}^{k})) \triangleq M_{\theta^{*}}(P(\textbf{x}, (x_i, y_i)_{i=1}^{k}))
     \label{LLM_out}
\end{equation}
for a model $M_{\theta^{*}}$ parameterized by $\theta^{*}$.

Given a few additional samples, we then optimize $\mathbf{A}$ and $\mathbf{b}$ using a validation set, starting with a zero initialization; we find that our method exhibits remarkable insensitivity to initialization and works quite well for zero/random initialization (see Figure \ref{table_init_comp}). More specifically, given a validation set $(x_i^{\rm v}, y_i^{\rm v})_{i=1}^{N_{\rm v}}$ of size $N_{\rm v}$, we create $N_{\rm v}$ validation prompts via:
\begin{align} 
    P_{i}^{\rm v} = P(x_i^{\rm v}, (x_j, y_j)_{j=1}^{k}).
    \label{Eq_Pv}
\end{align}
Our objective is to solve the following problem
\begin{align} 
     \min_{\mathbf{A}, \mathbf{b}} \,\,\sum_{i=1}^{N_{\rm v}} \mathcal{L}(\theta^{*}, \mathbf{A}, \mathbf{b}; P_{i}^{\rm v})
     \label{opt_prob}
\end{align}
where $\mathcal{L}$ represents the loss function of the calibration parameters $\mathbf{A}$ and $\mathbf{b}$.
We use a gradient-based optimizer to optimize $\mathbf{A}$ and $\mathbf{b}$ using prompts $P_{i}^{\rm v}$. Algorithm \ref{LinC} encapsulates our proposed LinC method. While we employ the stochastic gradient descent algorithm, our method can be utilized with any available optimization algorithm. LinC incurs negligible computational overhead and can be implemented in just a few lines of code to compute and store $\mathbf{A}$ and $\mathbf{b}$. {Moreover, LinC utilizes only $k+N_v$ extra samples to learn $\mathbf A$ and $\mathbf b$.} Finally, test predictions are obtained by calculating $\mathbf{A}\mathbf{p}+\mathbf{b}$ and taking the argument of the maxima after the softmax operator. 

\paragraph{Comparison with other methods.}
Unlike   calibration methods that rely on the raw data $(x_i^{\rm v}, y_i^{\rm v})_{i=1}^{N{\rm v}}$, our approach draws inspiration from ICL and employs prompts $\{P_{i}^{\rm v}\}_{i=1}^{N_{\rm v}}$ to learn the calibration parameters. Moreover, ICL performance is limited by the maximum input sequence length constraint of the underlying language model, causing poor scalability with an increase in the number of available training samples as shown in Figure \ref{max_seq_len}. In contrast, our approach is scalable with respect to the number of available data samples, since we use the available samples to optimize the calibration parameters. In fact, our method requires only a handful of additional samples (typically in the range 10-100) to effectively optimize the calibration parameters, {maintaining the few-shot regime} and making it much more sample efficient than fine-tuning LLMs which require orders of magnitude larger number of samples. {LinC is also much more computationally efficient as compared to other methods used to enhance LLM performance, such as fine-tuning, since it optimizes extremely low dimensional calibration parameters. Lastly, unlike in-context fine-tuning and meta-training methods \cite{min2021metaicl, wei2021finetuned}, LinC operates using API access alone and doesn't need access to the LLM architecture and weights.}
\begin{algorithm}[tb]
	\caption{Linear Probe Calibration ($\sf{LinC}$)}
	\label{LinC}
	\begin{algorithmic}[1]
        \State Input: LLM $\theta^{*}$, validation set $(x_i^{\rm v}, y_i^{\rm v})_{i=1}^{N_{\rm v}}$, $k$-shots $(x_j, y_j)_{j=1}^{k}$, loss function $\mathcal{L}$
        \State Initialize: parameters $\mathbf{A}_{0}^{0}, \mathbf{b}_{0}^{0}$ via zero initialization, step-size $\alpha$, number of epochs $T$
        \State For each $P_{i}^{\rm v}$ in \eqref{Eq_Pv}, obtain logits \eqref{LLM_out}: $\mathbf{p}_{i}^{\rm v} = \mathbf{p}(P_{i}^{\rm v})$
        \For{$t=1,\cdots, T$}
            \For{$i=1,\cdots, N_{\rm v}$}
                \State Compute  $\hat{\mathbf{p}}_{i}^{\rm v} = \mathbf{A}_{t-1}^{i-1}\mathbf{p}_{i}^{\rm v} + \mathbf{b}_{t-1}^{i-1}$
                \State $\mathbf{A}_{t-1}^{i} = \mathbf{A}_{t-1}^{i-1} - \alpha \nabla_{\mathbf{A}} \mathcal{L}(\hat{\mathbf{p}}_{i}^{\rm v}, y_i^{\rm v})$
                \State $\mathbf{b}_{t-1}^{i} = \mathbf{b}_{t-1}^{i-1} - \alpha \nabla_{\mathbf{b}} \mathcal{L}(\hat{\mathbf{p}}_{i}^{\rm v}, y_i^{\rm v})$
            \EndFor
            \State $\mathbf{A}_{t}^{0} = \mathbf{A}_{t-1}^{N_{\rm v}}$ and $\mathbf{b}_{t}^{0} = \mathbf{b}_{t-1}^{N_{\rm v}}$
        \EndFor
       \State return $\mathbf{A}_{T}^{0}, \mathbf{b}_{T}^{0}$
    \end{algorithmic} 
\end{algorithm}

\begin{table}[h]
\begin{center}
\small
  \begin{tabular}{ccllllllllll}
    \toprule
      Shot & Method & SST-2  &TREC  & Subj  \\
     
    \midrule
     \multirow{3}{*}{4-shot} & NoC & 66.3$_{13.1}$ & 24.0$_{6.3}$ & 52.8$_{4.9}$\\
     & ConC & 78.9$_{8.4}$ & 41.3$_{4.7}$ & 67.8$_{8.9}$\\
     & NoC* & 68.9$_{8.6}$ & 35.9$_{0.7}$ & 69.3$_{10.5}$ \\
     & ConC* & 75.9$_{3.9}$ &  {43.5}$_{2.5}$ & 61.9$_{10.2}$\\
     & LinC &  \textbf{{86.2}$_{2.9}$} &  \textbf{{45.9}$_{3.3}$} &  \textbf{{72.9}$_{11.1}$}\\
     
    \specialrule{0em}{2pt}{1pt}
    
     \multirow{3}{*}{8-shot} & NoC & 57.2$_{8.0}$ & 31.8$_{8.1}$ & 56.6$_{10.7}$ \\
     & ConC & 73.7$_{10.5}$ &  {45.4}$_{1.7}$ & 68.1$_{9.0}$\\
     & NoC* & 62.7$_{7.7}$ & 40.5$_{7.1}$ & 70.4$_{8.8}$ \\
     & ConC* & 75.8$_{8.1}$ &  {47.2}$_{2.3}$ & 63.0$_{7.7}$\\
     & LinC & \textbf{{79.1}$_{10.0}$} & \textbf{48.0$_{5.4}$} &  \textbf{{76.8}$_{3.5}$}\\
    \bottomrule
  \end{tabular}
  \caption{Comparison under same number of samples.}
  \label{table_same_samples}
\end{center}
\end{table}

\section{LinC: A Viable Solution} 
Before presenting our results, we first highlight the significance of linear calibration and why it is important when we are handicapped by limited resources. Therefore we ask: \emph{assuming that the maximum sequence length limit is not exceeded, what is the best way to use the additional samples in ICL? Is calibration the optimal way to use these additional samples?} 
To answer this question, we introduce two additional baselines, NoC* and ConC*\footnote{for details about NoC and ConC, see Section \ref{section_4.2}}, which utilize the same 10 validation samples used for training the calibration parameters in our method LinC. For these baselines, these 10 samples are treated as additional in-context test demonstrations within the test prompts.

Consequently, NoC* and ConC* should be considered to be in the 14-shot and 18-shot regimes instead of the 4-shot and 8-shot regimes, respectively. The choice of using 10 samples is to ensure that the maximum length limit does not exceed for any of the datasets used in this experiment. Table \ref{table_same_samples} shows the results on three different datasets in the 4-shot and 8-shot settings on GPT-2-XL. It is noteworthy that for all three datasets, using the same samples to learn the calibration parameters is better than using them within the test prompts as additional test demonstrations.

\begin{figure}[t]
\begin{center}
    \includegraphics[width=0.44\textwidth]{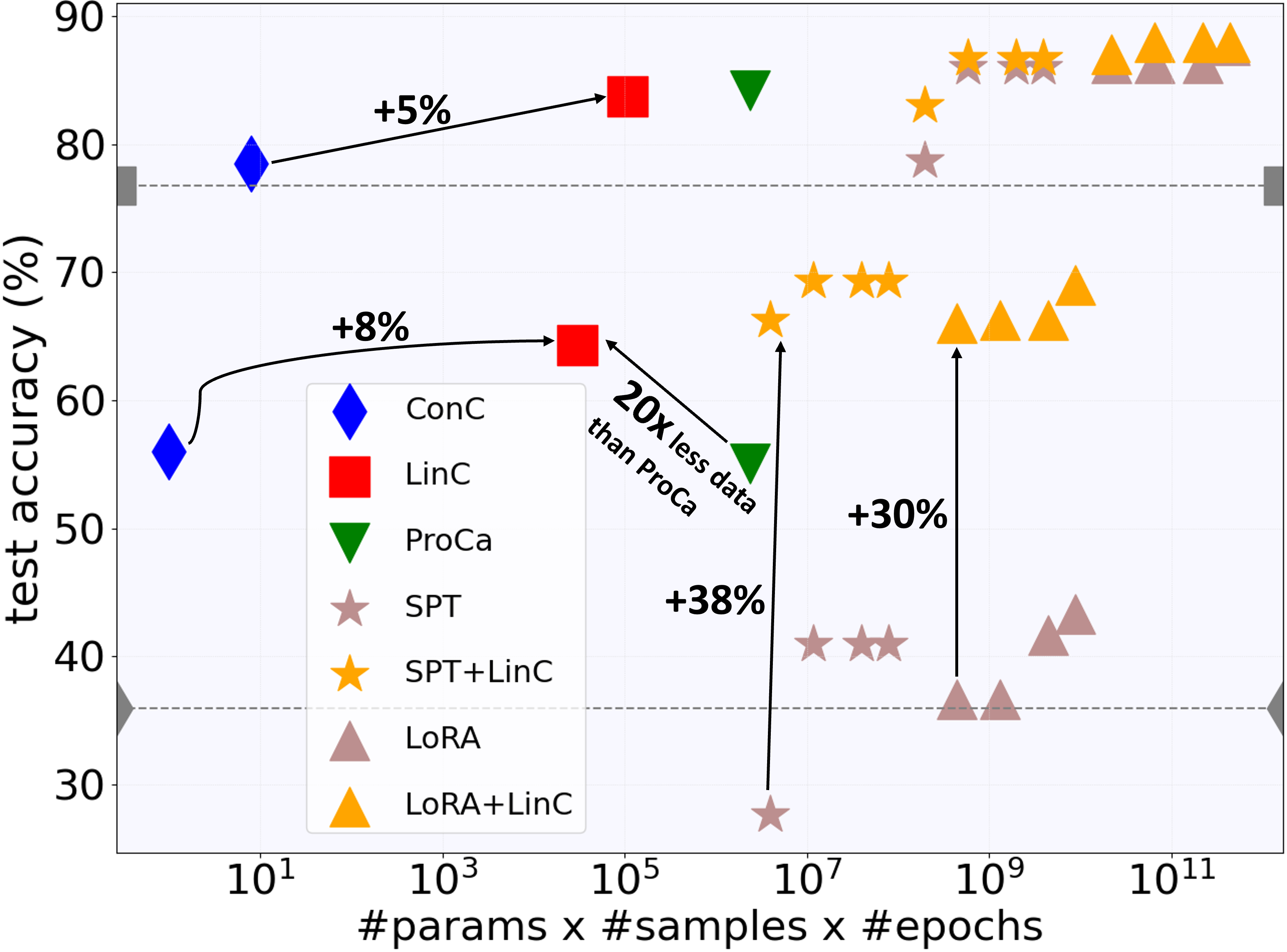}
\end{center}
\vspace{-0.1cm}
\caption{LinC outperforms ICL on all few-shot experiments, and substantially enhances PEFT, especially in the low resource regime, while maintaining almost identical data and compute requirements.}
  \label{ICL_params_samples_epoch}
\end{figure}

Moreover, we encapsulate our finding in Figure \ref{ICL_params_samples_epoch} by assessing the trade-off between performance and resource consumption, quantified as the product of trainable parameters, samples, and epochs. The raw ICL baselines (0/8-shot) are denoted by horizontal lines. Notably, LinC outperforms ICL baselines by a substantial margin while maintaining resource efficiency. In addition, it enhances the performance of Parameter-Efficient Fine-tuning (PEFT) methods, such as Soft Prompt Tuning (SPT) \cite{lester2021power} and Low-Rank Adaptation (LoRA) \cite{hu2021lora}, particularly in scenarios with limited data and computational resources. These observations render LinC particularly effective in situations where compute resources are limited and training data is scarce.

\begin{table}[h]
  \begin{center}
  \begin{tabular}{|c|c|c|c|c|c|c|c|c|c|c|}
    \hline
    \multirow{5}{0.1cm} \textbf{Dataset}  & \multicolumn{3}{c|}{\textbf{GPT-J (6B)}} \\
    \cline{2-4}
    &  \textbf{ICL} & \textbf{ConC} & \textbf{LinC}  \\
    \hline
    SST-2               & \underline{0.0592}& 0.1974  & \textbf{0.0458}  \\ \hline  \!\! \!\!   
    SST-5 \!\! \!\!     & 0.2254 & \textbf{0.0933} & \underline{0.0970}   \\ \hline
    AGNews              & 0.2858 &  \underline{0.0686}  & \textbf{0.0577}   \\ \hline
\!\! \!\!    
    TREC  \!\!  \!\!   & 0.2934 & \textbf{0.1108}  & \underline{0.1613}    \\ \hline
    DBPedia            & 0.2713 & \underline{0.2281} & \textbf{0.0502}   \\ \hline
    RTE                & \underline{0.0471} & 0.0825  & \textbf{0.0421}    \\ \hline
  \end{tabular}
  \caption{Expected Calibration Error (ECE) comparison between baselines and LinC (a model with perfect calibration would exhibit an ECE of 0).}
  \label{table_ECE}
  \end{center}
\end{table}

\begin{table}[h]
  \begin{center}
  \small
  \begin{tabular}{|c|c|c|c|c|c|c|c|c|c|c|}
    \hline
    \multirow{5}{0.1cm} \textbf{Dataset ($D/C$)}  & \multicolumn{3}{c|}{\textbf{GPT-J (6B)}}\\
    \cline{2-4}
    &  \textbf{ICL} & \textbf{ConC} & \textbf{LinC}  \\
    \hline
    Hamster (5/2)       & 55.6±8.3 & 53.3±9.4  & \textbf{60.0±14.4}  \\ \hline
  \!\! \!\!   Customers (8/2)  \!\! \!\!    & 67.4±0.5 & 54.9±2.3  & \textbf{68.6±0.5} \\ \hline
    Breast (7/2)        & 66.7±3.3 & 55.8±9.2  & \textbf{71.3±3.3} \\ \hline
\!\! \!\!    Spambase (57/2)  \!\!  \!\!   & 40.0±0.0 & 51.2±9.5  & \textbf{61.1±1.7}   \\ \hline
    TAE (5/3)           & 45.2±4.6 & 48.4±9.5  & \textbf{50.5±11.0}  \\ \hline
    Vehicle (18/4)      & 26.9±0.7 & 29.0±2.8  & \textbf{29.0±1.5}  \\ \hline
    LED (7/10)          & 16.3±10.4 & 27.3±6.9 & \textbf{28.0±7.1}  \\ \hline
  \end{tabular}
  \caption{Performance comparison between baselines and LinC on OpenML datasets; $D$ represents the number of features and $C$ represents the number of classes.}
  \label{table_OpenML}
  \end{center}
\end{table}

\section{Experiments}
 
This section will demonstrate the effectiveness of LinC on several benchmarks in the few-shot and PEFT settings. We run our experiments on GPT-2-XL \cite{radford2019languagegpt2} with 1.5B parameters, GPT-J \cite{wang2021gptj} with 6B parameters, and Llama-2 \cite{touvron2023llama} with 13B\footnote{Note that this is the largest model that can fit into our available GPU memory.} parameters on 2 NVIDIA GeForce RTX 3090 GPUs. We thoroughly study different label proportions, prompt templates, and demonstration permutations.

\subsection{Experimental Setup} \label{section_4.2}
We evaluate the effectiveness of our LinC method using seven widely used text-classification datasets: sentiment analysis using SST-2 \cite{sst2-socher2013recursive} and SST-5 \cite{sst2-socher2013recursive},  topic classification using the 4-way AGNews \cite{dbpedia-zhang2015character} and 14-way DBPedia \cite{dbpedia-zhang2015character}, 6-way question classification using TREC \cite{trec}, textual entailment using binary RTE \cite{rte-dagan2006pascal} from
SuperGLUE \cite{wang2019superglue}, and subjectivity classification using Subj \cite{subj-pang2004sentimental}. We also tested LinC on seven diverse non-text classification tasks using OpenML \cite{vanschoren2014openml} datasets with varying number of classes and features.

A fixed prompt format was utilized for each dataset, unless stated otherwise, which is demonstrated alongside examples in Appendix \ref{Appendix_prompt_templates}, Table \ref{table_prompt_openml} and Table \ref{table_fixed_prompt_templates}.

\begin{figure}[t]
\begin{center}
    \includegraphics[width=0.4\textwidth]{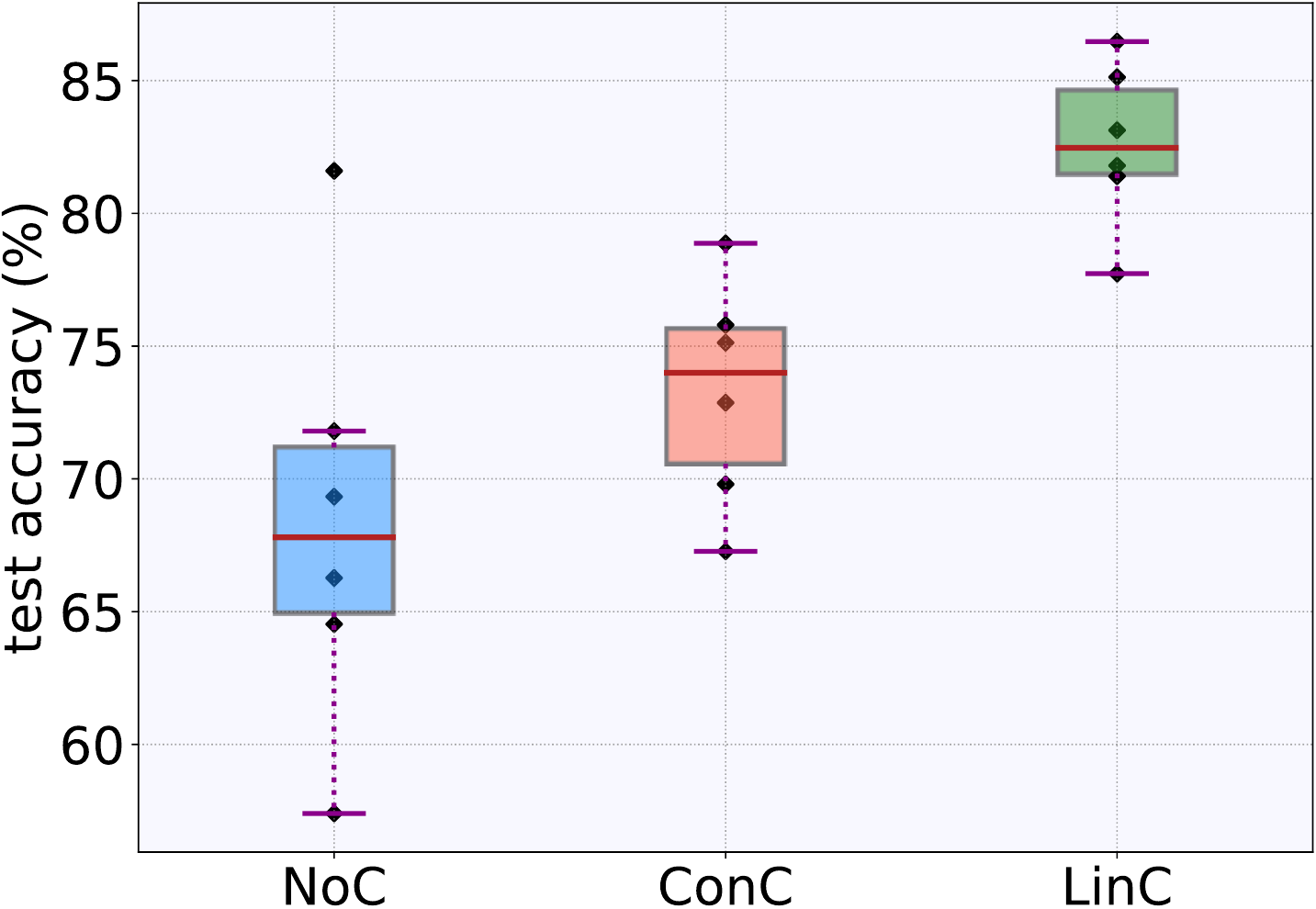}
\end{center}
\vspace{-0.1cm}
\caption{\small Comparison across six different templates.}
  \label{box_plot_template}
  \vspace{-0.15cm}
\end{figure}

\begin{table*}[h!]
\small
  \renewcommand{\arraystretch}{1.0}
  \setlength\tabcolsep{2.7pt}
  \centering
  \begin{tabular}{cclllllllllll}
    \toprule
      Model & Shots & Method & SST-2     & SST-5   & AGNews  &TREC &DBpedia & RTE  & Subj   & Avg   \\

    \midrule
    \specialrule{0em}{1pt}{1pt}
    \specialrule{0em}{0pt}{1pt}
    \midrule
     \multirow{12}{*}{GPT-2-XL 1.5B}  &
     \multirow{3}{*}{0-shot} & NoC &  64.5$_{0.0}$ & 33.7$_{0.0}$ & 44.3$_{0.0}$ & 28.7$_{0.0}$ & {58.7}$_{0.0}$ & 48.0$_{0.0}$ & {56.7$_{0.0}$} & {47.8} \\
     & & ConC & 70.9$_{0.0}$ & 20.3$_{0.0}$ &  {65.3}$_{0.0}$ & 41.7$_{0.0}$ & {50.0}$_{0.0}$ & 50.5$_{0.0}$ & {73.0$_{0.0}$} & {53.1}\\
     & & LinC &  \textbf{{71.6}$_{0.0}$} &  \textbf{{41.3}$_{0.0}$} & \textbf{65.7$_{0.0}$} &  \textbf{{42.0}$_{0.0}$} &  \textbf{{73.0}$_{0.0}$} &  \textbf{{54.5}$_{0.0}$} & \textbf{73.3$_{0.0}$} &  \textbf{60.2} \\
     
     \specialrule{0em}{2pt}{1pt}
     
     & \multirow{3}{*}{1-shot} & NoC & 59.7$_{13.2}$ & 28.3$_{9.7}$ & 39.6$_{10.3}$ & 27.1$_{5.9}$  & 40.5$_{15.9}$ & 53.4$_{1.0}$ & 54.6$_{8.8}$ & {43.3}\\
     & & ConC & 76.7$_{1.8}$ & 31.1$_{4.8}$ & 63.9$_{3.3}$ & 40.5$_{3.1}$ & 62.3$_{7.5}$ &  {52.5}$_{1.7}$ & 61.3$_{7.4}$ & {55.5}\\
     & & LinC &  \textbf{{83.2}$_{16.9}$} &  \textbf{{40.2}$_{3.7}$} &  \textbf{{64.8}$_{6.3}$} &  \textbf{{42.9}$_{5.1}$} &  \textbf{{63.1}$_{7.4}$} & \textbf{54.4$_{1.9}$} & \textbf{73.7$_{5.5}$} &  \textbf{60.3}\\
     
     \specialrule{0em}{2pt}{1pt}
     
    & \multirow{3}{*}{4-shot} & NoC & 66.3$_{13.1}$ & 34.1$_{4.8}$ & 40.4$_{14.1}$ & 24.0$_{6.3}$ & 66.7$_{10.2}$ & 52.2$_{3.2}$ & 52.8$_{4.9}$ & {48.1}\\
     & & ConC & 78.9$_{8.4}$ & 34.4$_{6.8}$ & 60.4$_{6.8}$ & 41.3$_{4.7}$ & 72.1$_{4.8}$ & 53.0$_{0.9}$ & 67.8$_{8.9}$ & {58.3}\\
     & & LinC &  \textbf{{87.1}$_{1.9}$} &  \textbf{{39.1}$_{5.6}$} &  \textbf{{69.0}$_{6.5}$} &  \textbf{{46.0}$_{3.2}$} &  \textbf{{73.1}$_{4.8}$} &  \textbf{{53.6}$_{0.6}$} & \textbf{73.6$_{10.8}$} &  \textbf{63.1}\\
     
    \specialrule{0em}{2pt}{1pt}
    
     & \multirow{3}{*}{8-shot} & NoC & 57.2$_{8.0}$ & 31.8$_{9.3}$ & 41.4$_{5.8}$ & 31.8$_{8.1}$ & 59.0$_{16.4}$ & 52.5$_{0.9}$ & {56.6$_{10.7}$} & {47.2}\\
     & & ConC & 73.7$_{10.5}$ &  {28.2}$_{5.4}$ & 57.9$_{12.2}$ & 45.4$_{1.7}$ & 71.8$_{5.7}$ & \textbf{53.4$_{1.1}$} & 68.1$_{9.0}$ & {56.9}\\
     & & LinC & \textbf{{79.1}$_{10.0}$} & \textbf{38.1$_{9.6}$} &  \textbf{{63.3}$_{7.2}$} &  \textbf{{48.1}$_{6.1}$} &  \textbf{{71.9}$_{5.5}$} &  {{53.2}$_{0.9}$} & \textbf{76.9$_{3.9}$} &  \textbf{61.5} \\

    \midrule
    \specialrule{0em}{1pt}{1pt}
    \specialrule{0em}{0pt}{1pt}
    \midrule
    
    \multirow{12}{*}{GPT-J 6B} &
    \multirow{3}{*}{0-shot} & NoC &  66.3$_{0.0}$ & 33.7$_{0.0}$ & 36.0$_{0.0}$ & 24.7$_{0.0}$ & 19.7$_{0.0}$ & 55.6$_{0.0}$ & {65.7$_{0.0}$} & {43.1} \\
     & & ConC & {58.0$_{0.0}$} & 40.7$_{0.0}$ &  {56.0}$_{0.0}$ & 40.0$_{0.0}$ & 48.0$_{0.0}$ & 52.4$_{0.0}$ & {58.7$_{0.0}$} & {50.5}\\
     & & LinC &  \textbf{{74.3}$_{0.0}$} &  \textbf{{46.0}$_{0.0}$} & \textbf{64.3$_{0.0}$} &  \textbf{{70.7}$_{0.0}$} &  \textbf{{69.3}$_{0.0}$} &  \textbf{{56.7}$_{0.0}$} & \textbf{70.7$_{0.0}$} &  \textbf{64.6} \\
     
     \specialrule{0em}{2pt}{1pt}
     
     & \multirow{3}{*}{1-shot} & NoC & 67.3$_{6.7}$ & 35.3$_{3.5}$ & 65.3$_{14.3}$ & 40.3$_{8.8}$  & 64.9$_{16.6}$ & 50.7$_{3.7}$ & 65.1$_{9.9}$ & {55.6}\\
     & & ConC & 88.3$_{1.7}$ & 46.9$_{3.1}$ & 74.9$_{5.8}$ & 62.6$_{4.7}$ & 80.2$_{3.3}$ &  {53.4}$_{1.2}$ & 59.1$_{2.2}$ & {66.5}\\
     & & LinC &  \textbf{{88.5}$_{1.7}$} &  \textbf{{50.4}$_{1.3}$} &  \textbf{{81.7}$_{4.6}$} &  \textbf{{63.0}$_{4.4}$} &  \textbf{{82.6}$_{3.7}$} & \textbf{56.0$_{1.8}$} & \textbf{69.9$_{7.5}$} &  \textbf{70.3}\\
     
     \specialrule{0em}{2pt}{1pt}
     
     & \multirow{3}{*}{4-shot} & NoC & 88.9$_{3.3}$ & 46.3$_{3.2}$ & 72.3$_{6.1}$ & 37.7$_{4.2}$ & 82.1$_{14.4}$ & 55.0$_{7.1}$ & 57.4$_{6.9}$ & {62.8}\\
     & & ConC & 92.8$_{3.1}$ & 49.4$_{4.8}$ & 75.2$_{4.1}$ & 46.0$_{5.1}$ & 88.9$_{4.9}$ & 56.0$_{1.9}$ & 65.3$_{11.8}$ & {67.7}\\
     & & LinC &  \textbf{{94.9}$_{0.9}$} &  \textbf{{51.1}$_{3.7}$} &  \textbf{{77.9}$_{5.5}$} &  \textbf{{65.3}$_{1.5}$} &  \textbf{{89.4}$_{5.4}$} &  \textbf{{58.4}$_{3.7}$} & \textbf{68.4$_{11.2}$} &  \textbf{72.2}\\
     
    \specialrule{0em}{2pt}{1pt}
    
     & \multirow{3}{*}{8-shot} & NoC & 91.8$_{6.0}$ & 44.5$_{4.3}$ & 76.8$_{9.9}$ & 43.5$_{6.3}$ & 88.6$_{3.2}$ & 60.3$_{2.8}$ & {82.2$_{5.7}$} & {69.7}\\
     & & ConC & 93.8$_{1.8}$ &  {44.4}$_{4.4}$ & 78.5$_{5.7}$ & 52.3$_{8.3}$ & \textbf{90.8$_{2.5}$} & 58.8$_{4.7}$ & 81.3$_{6.2}$ & {71.4}\\
     & & LinC  & \textbf{{94.8}$_{1.2}$} & \textbf{51.3$_{0.8}$} &  \textbf{{83.7}$_{1.8}$} &  \textbf{{67.3}$_{3.8}$} &  {{90.1}$_{3.9}$} &  \textbf{{63.2}$_{4.2}$} & \textbf{84.7$_{4.9}$}  &  \textbf{76.4} \\

    \midrule
    \specialrule{0em}{1pt}{1pt}
    \specialrule{0em}{0pt}{1pt}
    \midrule
    \multirow{12}{*}{Llama-2 13B} &
    \multirow{3}{*}{0-shot} &  NoC &  56.3$_{0.0}$ & 34.0$_{0.0}$ & 73.3$_{0.0}$ & 48.7$_{0.0}$ & 54.7$_{0.0}$ & 66.1$_{0.0}$ & {47.3$_{0.0}$} & {54.3} \\
    & & ConC & {69.3$_{0.0}$} & 33.3$_{0.0}$ &  {72.3}$_{0.0}$ & 71.3$_{0.0}$ & 75.0$_{0.0}$ & 67.5$_{0.0}$ & {47.0$_{0.0}$} & {62.2}\\
    & & LinC &  \textbf{{75.3}$_{0.0}$} &  \textbf{{47.3}$_{0.0}$} & \textbf{85.7$_{0.0}$} &  \textbf{{75.0}$_{0.0}$} &  \textbf{{90.7}$_{0.0}$} &  \textbf{{69.0}$_{0.0}$} & \textbf{48.3$_{0.0}$} &  \textbf{70.2} \\
     
     \specialrule{0em}{2pt}{1pt}
     
     & \multirow{3}{*}{1-shot} & NoC & 73.9$_{12.8}$ & 43.9$_{3.5}$ & 81.5$_{2.7}$ & 67.0$_{7.2}$  & 92.3$_{1.7}$ & 70.3$_{3.5}$ & 50.5$_{3.9}$ & {68.5}\\
     & & ConC &  \textbf{{94.1}$_{1.7}$} & 42.4$_{3.9}$ & 80.6$_{1.9}$ & 76.2$_{2.5}$ & 92.3$_{1.2}$ &  {60.9}$_{7.7}$ & 53.2$_{12.2}$ & {71.4}\\
     & & LinC &  \textbf{{94.1}$_{1.7}$} &  \textbf{{51.0}$_{1.6}$} &  \textbf{{84.1}$_{1.7}$} &  \textbf{{77.3}$_{2.6}$} &  \textbf{{93.6}$_{1.6}$} & \textbf{75.9$_{2.6}$} & \textbf{55.6$_{10.2}$} &  \textbf{75.9}\\
     
     \specialrule{0em}{2pt}{1pt}
     
     & \multirow{3}{*}{4-shot} & NoC & 92.9$_{2.6}$ & 48.7$_{4.1}$ & 82.7$_{3.7}$ & 62.5$_{14.7}$ & 94.2$_{1.0}$ & 68.8$_{8.7}$ & 72.0$_{11.8}$ & {74.5}\\
     & & ConC & 97.4$_{0.4}$ & 44.9$_{5.2}$ & 80.5$_{2.3}$ & 75.3$_{7.2}$ & 94.8$_{1.2}$ & 75.1$_{2.0}$ & 72.5$_{9.8}$ & {77.2}\\
     & & LinC &  \textbf{{97.5}$_{0.5}$} &  \textbf{{53.0}$_{0.9}$} &  \textbf{{86.0}$_{1.2}$} &  \textbf{{75.7}$_{2.1}$} &  \textbf{{95.3}$_{0.9}$} &  \textbf{{77.0}$_{0.9}$} & \textbf{78.7$_{9.6}$} &  \textbf{80.5}\\
     
    \specialrule{0em}{2pt}{1pt}
    
     & \multirow{3}{*}{8-shot} & NoC & 92.2$_{3.1}$ & 49.1$_{7.3}$ & 87.0$_{0.5}$ & 77.3$_{4.7}$ & 94.9$_{1.7}$ & 72.9$_{5.6}$ & \textbf{82.7$_{7.4}$} & {79.4} \\
     & & ConC & 96.7$_{0.5}$ &  {47.1}$_{4.4}$ & 83.8$_{1.6}$ & 79.1$_{3.7}$ & 94.6$_{1.9}$ & {75.3$_{3.0}$} & 82.5$_{5.5}$ & {79.9}\\
     & & LinC  & \textbf{{97.0}$_{0.2}$} & \textbf{51.4$_{6.1}$} &  \textbf{{87.1}$_{0.6}$} &  \textbf{{79.7}$_{3.9}$} &  \textbf{{95.0}$_{1.4}$} &  \textbf{{76.4}$_{0.6}$} & {82.1$_{3.1}$} &  \textbf{81.2} \\
    \bottomrule
  \end{tabular}
  
  \setlength{\abovecaptionskip}{0.4cm} 
  \caption{Comparisons among the conventional approach (NoC; \cite{GPT3}), ConC \cite{zhao2021calibrate} and LinC (\textit{Ours}) on GPT-2, GPT-J and Llama-2. We report the mean and the standard deviation of test accuracy across different
choices of the test demonstrations (the prompt   is fixed). We also report the average performance across seven datasets.}
  \label{table_GPTJ}
\end{table*}
 
Experiments were conducted under 0-shot, 1-shot, 4-shot and 8-shot learning settings. Five different sets of test demonstrations were chosen at random, and arranged in an arbitrary order in the prompt, and the mean and standard deviation were computed across all five test prompts. We used the cross-entropy loss as 
\begin{equation}
    \mathcal{L}(\theta^{*}, \mathbf{A}, \mathbf{b}; P_{i}^{\rm v}, y_{i}^{{\rm v}}) = - \sum_{c=1}^{C} y_{i,c}^{{\rm v}} \log(\Tilde{\mathbf{p}}_{c})
\end{equation}
 where $y_{i,c}^{\rm v}$ is the binary variable  indicating if class $c$ is the correct label for validation input $x_i^{\rm{v}}$, and $\Tilde{\mathbf{p}}_{c}$ denotes the output from the softmax defined in \eqref{eq3} with $\mathbf{p} = \mathbf{p}(P_{i}^{\rm v}) = \mathbf{p}(P(x_i^{\rm v}, (x_j, y_j)_{j=1}^{k}))$ defined in \eqref{LLM_out}. In \eqref{opt_prob}, we keep the loss general as we can use different losses depending on the task.

For each experiment, we fine-tuned the step size $\alpha$. The number of epochs $T$ was chosen from  $\{1, 5, 15, 50, 100\}$ and the number of validation prompts $N_{\rm v}$ was chosen from $\{1, 5, 10, 30, 100, 300\}$ (for details, see Appendix \ref{Appendix_experiments}). If the validation set is provided for a dataset, we utilize it as is. However, if the validation set is not available, we create one by randomly selecting a subset from the training set. As baselines, we used the vanilla ICL method \cite{GPT3} that does not use any calibration (NoC) and contextual calibration (ConC) \cite{zhao2021calibrate}. The results of ConC were replicated using the released code\footnote{\url{www.github.com/tonyzhaozh/few-shot-learning}}. 
Both ConC and our approach, LinC, utilize an affine transformation. However, the crucial distinction lies in the fact that \textbf{LinC acquires the transformation parameters through the learning process with a few additional samples} (following the same format as instruction-tuning). In contrast, ConC \textbf{does not engage in learning} and opts for a pre-defined initialization instead (for details, see Appendix \ref{LinC_vs_ConC}).
The demonstrations' labels were not artificially balanced. Moreover, we observed that our method performs well with any set of $k$-shot demonstrations used in the validation prompts, but utilizing a different set can lead to improved performance, thus we conducted experiments with different random seeds to obtain a better set of the validation demonstrations.

\subsection{Simulation Results} \label{section_results}
\textbf{LinC enhances both the average and minimum accuracy.}
Table \ref{table_GPTJ} shows the results on GPT-2-XL, GPT-J and Llama-2, respectively. LinC consistently outperforms baselines in almost all experiments, demonstrating its strong generalization ability across different model sizes and few-shot settings. 
We observe that, on average, LinC achieves up to $21 \%$ improvement as compared to the vanilla ICL baseline and $14 \%$ improvement as compared to contextual calibration for 0-shot learning on GPT-J. Moreover, in certain cases, LinC can deliver a significant boost in performance, up to $50 \%$ absolute improvement, as observed in the GPT-J 0-shot experiment on the DBPedia dataset. LinC demonstrates significant performance gain on some datasets, such as TREC, while moderate improvement is observed for other datasets, such as SST-2. 
Additionally, the performance improvement of LinC is more prominent on GPT-J than on GPT-2-XL and Llama-2. {We also compare our results with prototypical calibration \cite{han2022prototypical}, despite the fact that they use orders of magnitude larger number of samples and thus fall outside the few-shot learning regime. 

Table \ref{table_GPTJ-Proca} in Appendix \ref{Appendix_experiments} shows that LinC outperforms prototypical calibration in 16 out of 28 cases while using fewer number of samples (see Table \ref{table_samples_Proca}). When using the same number of samples, LinC outperforms prototypical calibration in 20 of 28 cases.}
{LinC's exceptional ability to generalize effectively could be supported by a recent discovery indicating that ICL in a conventional Transformer block is equivalent to adjusting the output layer using linear modeling of meta-learned deep data representations in few-shot settings, and LinC can be seen as explicitly adjusting this output layer with the additional samples \cite{von2022transformers}.}

 We also performed an experiment to show the impact of the model sizes (e.g., 7B vs 13B) on performance within the same model family (e.g., Llama-2); see Table \ref{table_Llama7B}. The results demonstrate that LinC consistently improves performance regardless of the model size. 

\textbf{LinC reduces variance across demonstrations.} Figure \ref{box_plot_std} shows the difference in standard deviation between the calibration methods and the NoC baseline for all GPT-J experiments in Table \ref{table_GPTJ}. In most cases, LinC significantly reduces variance while only slightly increasing it in the remaining cases. Moreover, on average, LinC achieves a greater reduction in standard deviation than ConC, indicating that the predictions made by LinC are more consistent and reliable.

\textbf{LinC improves the Expected Calibration Error (ECE).} To further evaluate LinC's model calibration, we use the ECE metric \cite{Naeini2015} that is widely-used to quantify the alignment between predicted and actual probabilities. Table \ref{table_ECE} shows the results on GPT-J, 0-shot setting (see Appendix \ref{Appendix_experiments} for other settings). In most instances, LinC demonstrates the lowest ECE, and in others, it ranks as the second-best method, with only a minimal difference from the best method.

\textbf{{LinC improves accuracy and reduces variance across varying prompt templates}.} Next  we keep the set of demonstrations fixed and vary the prompt format. We use six different prompt formats and label spaces for SST-2 dataset (for details, refer to Appendix \ref{Appendix_prompt_templates}, Table \ref{table_prompt_sst2}) on GPT-2-XL under 4-shot setting. From Figure \ref{box_plot_template}, we observe that while ConC generally enhances the average accuracy, it results in high variance. In contrast, LinC exhibits a considerable enhancement in accuracy with much lower variance, demonstrating its effectiveness in improving the model's performance across various prompt templates.

\begin{figure}[t]
    \centering
     \includegraphics[width=0.5\textwidth]{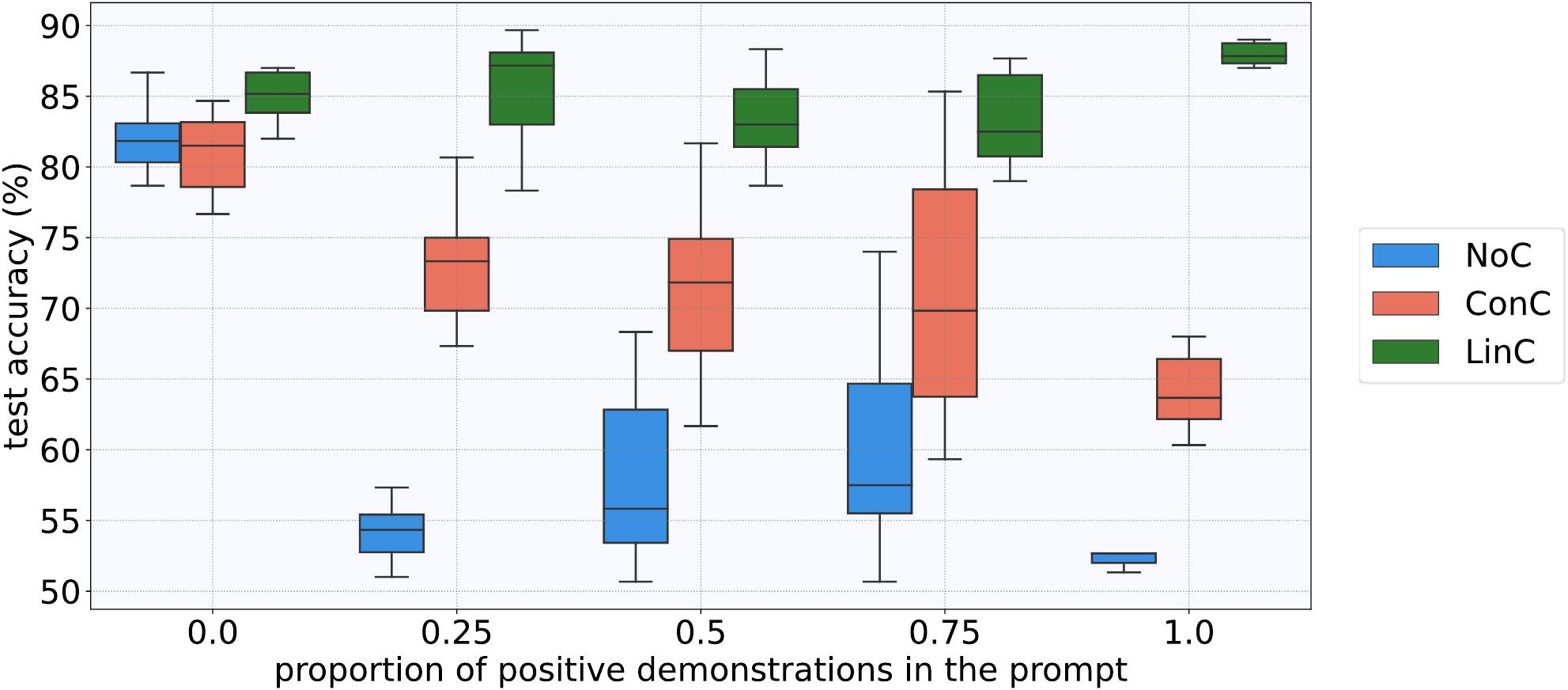}
  \vspace{-0.2cm}
    \caption{Comparison under varying label proportions and permutations of demonstrations; each box represents the test accuracy obtained from eight random permutations.}
    \label{box_plot_label_prop}
\end{figure}

\textbf{LinC is robust to class imbalance and permutations of demonstrations.} Prior works \cite{zhao2021calibrate, lu2021fantastically} have shown that the order in which the demonstrations are set in the prompt can significantly affect the performance.
We evaluated the performance of our method on five 8-shot prompts for SST-2 on GPT-2-XL, each with varying class proportions, and measured the accuracy across eight random permutations for each proportion. The demonstrations in the test prompt are kept the same for each proportion for both baselines and our method. Figure \ref{box_plot_label_prop} illustrates that vanilla ICL (NoC) can achieve satisfactory performance with certain proportions and permutations, but performs much worse in most cases. This result is in line with the observations made in previous works \cite{lu2021fantastically}. Moreover, ConC achieves superior test accuracy on average when compared to NoC but the performance is volatile across different permutations. LinC stands out as the most effective model and displays  low variance across different permutations, suggesting that it is robust to varying class proportions and permutations.
\begin{figure}[t]
\begin{center}
    \includegraphics[width=0.38\textwidth]{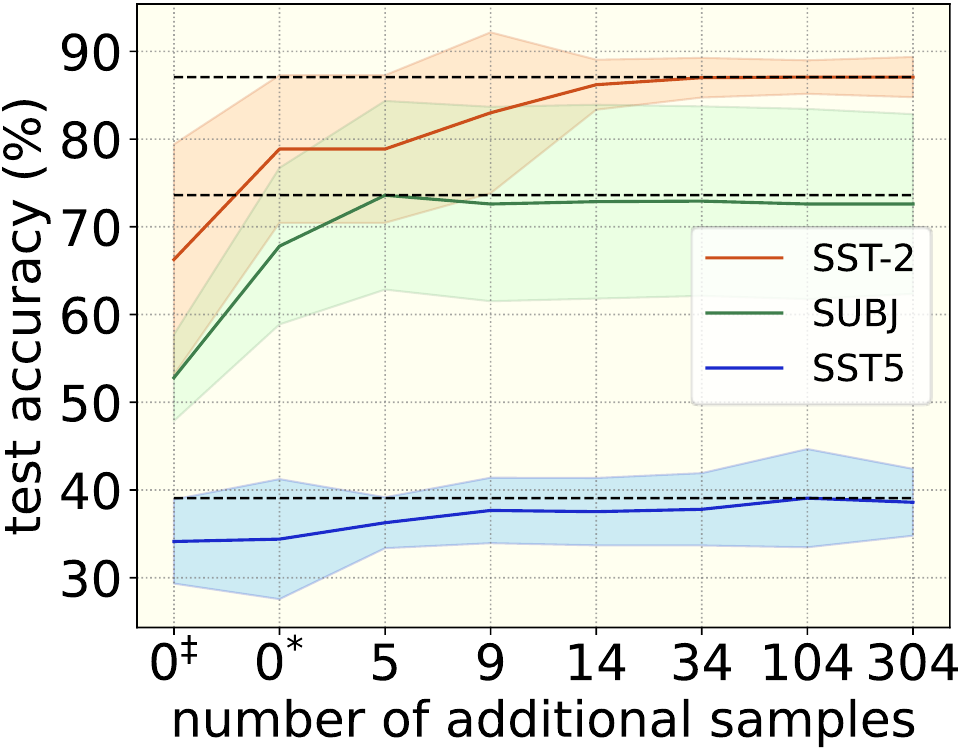}
  \end{center}
  \caption{\small Performance across varying validation sizes; $^\ddagger$ denotes NoC, and $^{*}$ denotes ConC. Black dotted line marks the maximum accuracy.}
  \label{ablation_samples}
  \vspace{-0.2cm}
\end{figure}

\textbf{LinC needs very few additional samples for improving ICL.} Figure \ref{ablation_samples} investigates the effect of additionally available validation samples on the performance of three different datasets on GPT-2-XL under the 4-shot setting. The number of extra samples employed in the validation prompts (i.e. $N_{\rm v} + k$) is plotted along the x-axis.
We observe that performance can be greatly improved by increasing the number of validation samples within a certain small range. However, further increasing the number of samples does not yield any additional improvement. LinC demonstrates high sample efficiency, achieving maximum accuracy with less than 30 additional samples on most datasets. Remarkably, some datasets require only five additional samples (e.g., Subj) to achieve maximum accuracy.

\textbf{LinC also improves performance on non-language tasks.} We also evaluated the performance of our method on non-language classification tasks using seven real tabular datasets in OpenML \cite{vanschoren2014openml} following the settings in \cite{dinh2022lift}. Table \ref{table_OpenML} demonstrates the performance improvement of LinC over raw ICL and ConC across all tabular datasets, regardless of the number of classes (C) and data features (D). In contrast, ConC sometimes falls short in improving performance, especially in datasets with fewer classes and features, and has worse performance than raw ICL in datasets such as Hamster, Customers, and Breast.

\textbf{LinC improves PEFT methods in the low-data low-compute regime.} We evaluated the performance of LinC on two popular PEFT methods: 1) SPT \cite{lester2021power} and 2) LoRA \cite{hu2021lora} (details in Appendix \ref{PEFT_settings}). Tables \ref{table_SPT} and \ref{table_LoRA} demonstrate that LinC consistently boosts both methods during fine-tuning across different sample sizes. However, performance improvement is most prominent when data and compute resources are limited. For example, for a sample size of 120 (out of 120k available training samples), LinC yields an accuracy improvement of $+38.7\%$ for SPT and $+29.3\%$ for LoRA. This makes LinC especially viable in scenarios with constrained compute and a scarcity of data.

\section{Related Work}
\textbf{Understanding ICL.}
Some recent works focus on explaining the working of ICL. For example, \cite{xie2021explanation} suggested that ICL is an implicit Bayesian inference and proved it through a synthetic dataset with a mixture of hidden Markov models in pretraining. \cite{garg2022can} showed that Transformers can learn effective learning algorithms for unseen linear functions based on demonstration samples and achieve comparable error to least squares estimator in ICL models. \cite{dai2022can} explain that ICL can be considered as implicit finetuning, where LLM generates meta-gradients from in-context demonstrations to adjust the model's behavior. \cite{li2023transformers} framed ICL as an algorithm learning problem and demonstrated that Transformers can effectively implement a function class through implicit empirical risk minimization based on demonstrations. \cite{von2022transformers} showed self-attention-only Transformers trained on simple regression tasks exhibit significant similarity to models learned by gradient descent, revealing how trained Transformers execute gradient descent during their forward pass. \cite{chan2022data} showed that ICL's performance is influenced by the distributional properties of training data, with improved performance observed when training data consists of clustered examples and sufficient rare classes. ICL is also inherently connected to multi-task learning \cite{caruana1997multitask,zhang2021survey} and meta-learning \cite{finn2017maml, abbas2022, chen2022meta, Kang_2023_CVPR, 10.5555/3586589.3586618}. But a key difference between ICL and those methods is that, in
ICL, adaptation to a new task is done implicitly through input prompt not running explicit gradient-based optimization. 
 
\textbf{Enhancing ICL.} To enhance the performance of ICL, meta-learning has been introduced by \cite{chen2021meta, min2021metaicl} to improve the adaptation of LLMs to ICL. \cite{wei2022chain} suggest augmenting the demonstrations by incorporating human-aided reasoning steps, leading to an improvement in performance on a range of arithmetic and reasoning tasks. \cite{wei2021finetuned, sanh2021multitask} suggest instruction tuning as a method to further pre-train the language model with a variety of downstream tasks in a shared prompting format. Several works focus on finding good in-context demonstrations to improve ICL performance \cite{liu2021makes, levy2022diverse}. \cite{wang2023large} investigates ICL using a Bayesian approach,  and proposes an algorithm for selecting optimal demonstrations, showing empirical improvement compared to a random selection baseline.
\cite{si2022prompting} studies model reliability using four different facets. 
\cite{han2022prototypical} proposes estimating prototypical clusters for all classes, mapping each cluster to the corresponding label, and calibrating the test predictions by their most likely cluster. 
It has been found in  \cite{jiang2023generative} that the sensitivity of the language model stems from the label shift of the model in the data distribution, where the model exhibits a shift in the label marginal while maintaining a strong label conditional. Their solution involves a generative calibration approach, adjusting the label marginal through Monte-Carlo sampling over the in-context model to calibrate the predictive distribution. \cite{zhou2023batch} introduce Batch Calibration (BC), a zero-shot calibration method aimed at reducing bias from the batch.
Closely related to our work is \cite{zhao2021calibrate} which observed that the few-shot performance of language models is not consistent across different in-context settings. Additionally, \cite{zhao2021calibrate} notes that language models tend to predict certain labels due to bias or demonstration permutations. However, the proposed calibration method of selecting content-free test inputs in \cite{zhao2021calibrate} cannot accurately reflect the bias of models, which can result in sub-optimal performance. In contrast, our LinC approach achieves calibration of bias by accurately optimizing calibration parameters at the expense of minimal compute and data. 
              
\section{Conclusions}
 
This paper investigates in-context learning (ICL), which relies on GPT-like models to generate outputs based on in-context demonstrations. Our findings reveal that ICL predictions may be unreliable when evaluated using Shannon entropy. To overcome these limitations, we propose the linear probe calibration (LinC) method, which significantly boosts the test performance of GPT models on various benchmark datasets with only a minimal number of additional samples. LinC's ability to reduce ECE and variance across different sets of demonstrations, and maintain robustness towards varying label proportions, prompt templates, and demonstration permutations implies that the predictions made by the LLM were more reliable, supporting our original conclusion from the Shannon entropy metric analysis. 
We believe that these findings carry important implications for future research and the development of more reliable and effective natural language processing models.

\section*{Limitations and Future Work}
While our focus remains on calibrating the model for better accuracy and reliability, it is worth discussing how to combine our framework with approaches for selecting better examples and prompt templates, such as those proposed in \cite{liu2021makes, sorensen2022information}. Our work focuses on querying LLMs, which can generate content with potential ethical risks such as fairness and bias; thus, combining our framework with methods \cite{gupta2022mitigating} that mitigate such risks is worth further discussion. One limitation of our work is the choice of $k$-shot validation demonstrations used to optimize the calibration parameters can impact the quality of the learned parameters, potentially leading to suboptimal results. Besides overcoming this limitation, we aim to expand our method to other NLP tasks such as summarization, text generation, and generative question answering.

 \section*{Acknowledgment}
The work of T. Chen and M. Abbas was supported by the Rensselaer-IBM AI Research Collaboration (\url{http://airc.rpi.edu}), part of the IBM AI Horizons Network (\url{http://ibm.biz/AIHorizons}). 

\bibliography{refer}
\bibliographystyle{plain}

\onecolumn
\aistatstitle{Supplementary Material for\\ ``Enhancing In-context Learning via Linear Probe Calibration"}
\appendix

\section{Prompt Templates} \label{Appendix_prompt_templates}
\begin{table}[h]
\centering
\mycfs{8.2}
\aboverulesep = 0.205mm
\belowrulesep = 0.405mm
\setlength{\tabcolsep}{2pt}
\begin{tabular}{cp{8cm}p{2.5cm}}
\toprule
\textbf{Format \#} & \textbf{Prompt Template} & \textbf{Label Space} \\
\midrule
1 & Review: Perhaps the best sports movie I have ever seen. \newline\vspace{2.6pt}\hspace{-0.1cm} Sentiment: Positive \newline Review: This pathetic junk is barely an hour long. \newline Sentiment: &  Positive, Negative \\

\midrule
2 & Input: Perhaps the best sports movie I have ever seen. \newline\vspace{2.6pt}\hspace{-0.1cm} Prediction: Positive \newline Input: This pathetic junk is barely an hour long. \newline Prediction: &  Positive, Negative \\

\midrule
3 & Review: Perhaps the best sports movie I have ever seen. \newline\vspace{2.6pt}\hspace{-0.1cm} Sentiment: good \newline Review: This pathetic junk is barely an hour long. \newline Sentiment: &  good, bad \\

\midrule
4 & Input: Perhaps the best sports movie I have ever seen. \newline\vspace{2.6pt}\hspace{-0.1cm} Prediction: good \newline Input: This pathetic junk is barely an hour long. \newline Prediction: &  good, bad \\

\midrule
5 & Perhaps the best sports movie I have ever seen. \emph{My overall feeling was that the movie was} \underline{good} \newline \newline This pathetic junk is barely an hour long. \emph{My overall feeling was that the movie was} \rule{0.6cm}{0.15mm} & good, bad \\

\midrule
6 & Review: Perhaps the best sports movie I have ever seen. \newline
Question: Is the sentiment of the above review Positive or Negative? \newline\vspace{2.6pt}\hspace{-0.1cm}
Answer: Positive \newline Review: This pathetic junk is barely an hour long. \newline
Question:  Is the sentiment of the above review Positive or Negative? \newline
Answer: & Positive, Negative \\

\bottomrule
\end{tabular}
\vspace{0.1cm}
\caption{\small A list of different prompt templates that were used to investigate the impact of templates on SST-2. For brevity, here we show only one demonstration.}
\label{table_prompt_sst2}
\aboverulesep = 0.605mm
\belowrulesep = 0.984mm
\end{table}  

\begin{table}[h]
\centering
\mycfs{8.2}
\aboverulesep = 0.205mm
\belowrulesep = 0.405mm
\setlength{\tabcolsep}{2pt}
\begin{tabular}{cp{8cm}p{2.5cm}}
\toprule
\textbf{Dataset} & \textbf{Prompt Template} & \textbf{Label Space} \\
\midrule
All OpenML datasets & When we have x1=\textcolor{red}{r.x1}, x2=\textcolor{red}{r.x2}, $\dots$, xK=\textcolor{red}{r.xD}, what should be y? $\#\#\#$ y=\textcolor{red}{r.y} $@@@$  &  ~~0,\dots,C \\

\bottomrule
\end{tabular}
\vspace{0.1cm}
\caption{\small Prompt template used for non-language classification tasks using the real tabular datasets in OpenML \cite{vanschoren2014openml}; for a sample $r$, $D$ denotes the number of features;  Following \cite{dinh2022lift}, we follow OpenAI's recommendation by using "$\#\#\#$" for separating questions and answers, and "$@@@$" to indicate the end of answer.}
\label{table_prompt_openml}
\aboverulesep = 0.605mm
\belowrulesep = 0.984mm
\end{table}

\begin{table}[H]
\centering
\mycfs{8.5}
\begin{tabular}{p{1.0cm}p{8cm}p{4cm}}
\toprule
\textbf{Dataset} & \textbf{Prompt Template} & \textbf{Label Space} \\
\midrule
{SST-2} & Review: Perhaps the best sports movie I have ever seen. \newline Sentiment: Positive
\vspace{3.9pt}\hspace{-0.1cm}

Review: This pathetic junk is barely an hour long. \newline Sentiment: &  Positive, Negative \\

\midrule
AGNews & Classify the news articles into the categories of World, Sports, Business, and Technology. \vspace{3.9pt}\hspace{-0.1cm}

Article: UK lender Barclays says it is in talks with South Africa's Absa about buying a majority stake in the bank. \newline Answer: Business
\vspace{3.9pt}\hspace{-0.1cm}

Article: New music sharing network allows users to amass points by referring buyers. \newline Answer: &  World, Sports, Business, Technology \\

\midrule
TREC & Classify the questions based on whether their answer type is a Number, Location, Person, Description, Entity, or Abbreviation. \vspace{3.9pt}\hspace{-0.1cm}

Question: What does the abbreviation AIDS stand for? \newline
Answer Type: Abbreviation
\vspace{3.9pt}\hspace{-0.1cm}

Question: What country do the Galapagos Islands belong to? \newline
Answer Type: &  Number, Location, Person, Description, Entity, Abbreviation \\
\midrule
DBPedia & Classify the documents based on whether they are about a Company, School, Artist, Athlete, Politician, Transportation, Building, Nature, Village, Animal, Plant, Album, Film, or Book.
\vspace{3.9pt}\hspace{-0.1cm}

Article: Hoodlum \& Son is a 2003 comedy-crime film. \newline
Answer: Film
\vspace{3.9pt}\hspace{-0.1cm}

Article:  Nachan Main Audhay Naal is the seventh album of Pakistani pop and bhangra singer Abrar-ul-Haq. It was released on March 2007. \newline
Answer: &  Company, School, Artist, Athlete, Politician, Transportation, Building, Nature, Village, Animal, Plant, Album, Film, Book \\

\midrule
SST-5 & Review: The film is bright and flashy in all the right ways. \newline
Sentiment: great
\vspace{3.9pt}\hspace{-0.1cm}

Review: The film never finds its tone and several scenes run too long. \newline
Sentiment: & terrible, bad, okay, good, great \\

\midrule
Subj & Input: All social structures break down and a new world order emerges from the heart of the desert. \newline
Type: objective
\vspace{3.9pt}\hspace{-0.1cm}

Input: A zombie movie in every sense of the word -- mindless, lifeless, meandering, loud , painful, obnoxious. \newline
Type: & objective, subjective \\

\midrule
RTE & Experts say that Mr. Abbas will need that big win to show that he has the support of most Palestinian people in order to push through his aims of peace talks with Israel.\newline
question: Analysts had said that Mr. Abbas needed a large margin of victory in order to push his agenda of peace talks with Israel. True or False?\newline
answer: True
\vspace{3.9pt}\hspace{-0.1cm}

The city is twinned with Glasgow, Dortmund, Pleven, and Le Mans. \newline
question: Dortmund is twinned with Glasgow. True or False?\newline
answer: & True, False \\
\bottomrule
\end{tabular}
\vspace{0.2cm}
\caption{\small The prompts templates used for different datasets. For brevity, here we show only one demonstration per dataset.}
\label{table_fixed_prompt_templates}
\end{table}

\clearpage
\section{Additional Experiments} \label{Appendix_experiments}
In this section, we provide details of the experimental set-up and present additional results.

 \begin{figure*}[h!]
    \centering
 \vspace{-0.0cm}
    \includegraphics[width=1.0\textwidth]{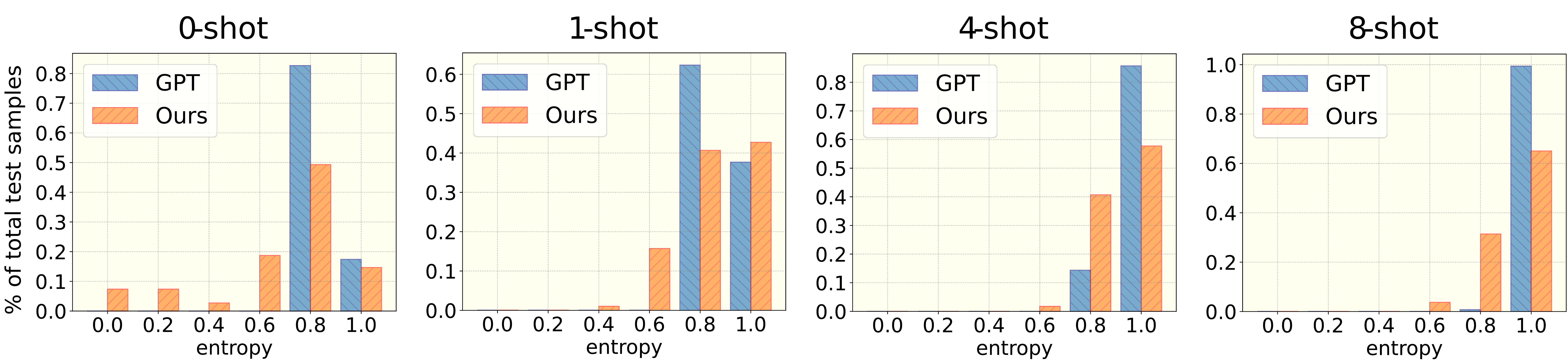}
    \vspace{-0.1cm}
    \caption{\small Shannon entropy histograms of using vanilla ICL on GPT-2-XL (1.5B) vs our method on Subj (higher entropy implies higher uncertainty); we use logarithmic base two. Refer to Section~\ref{sec:shannon} for a detailed explanation.}
    \vspace{-0.4cm}
\end{figure*}
\begin{figure*}[h!]
    \centering
    \includegraphics[width=1.0\textwidth]{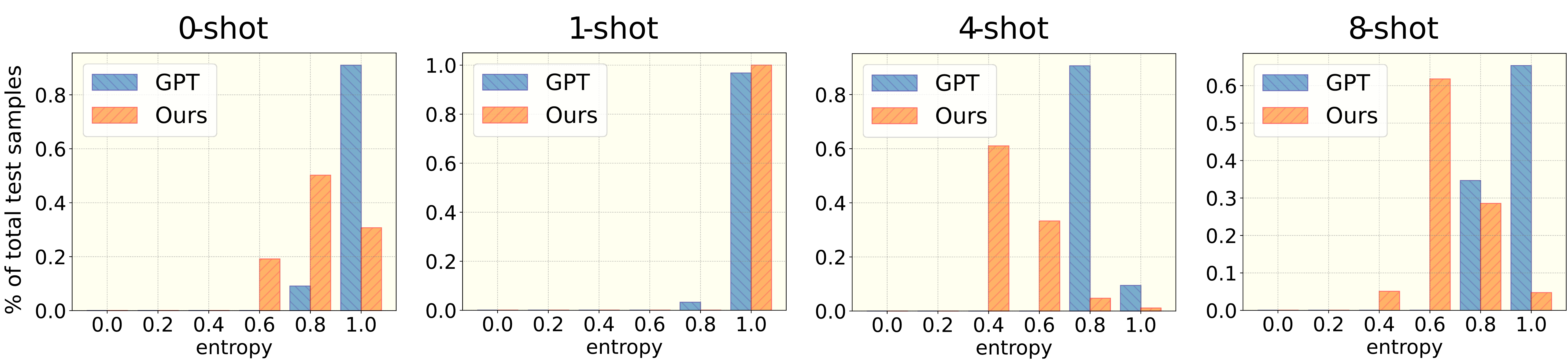}
    \vspace{-0.1cm}
    \caption{\small Shannon entropy histograms of using vanilla ICL on GPT-2-XL (1.5B) vs our method on RTE (higher entropy implies higher uncertainty); we use logarithmic base two. Refer to Section~\ref{sec:shannon} for a detailed explanation.}
    \vspace{-0.1cm}
\end{figure*}
\begin{figure*}[h!]
    \centering
    \includegraphics[width=1.0\textwidth]{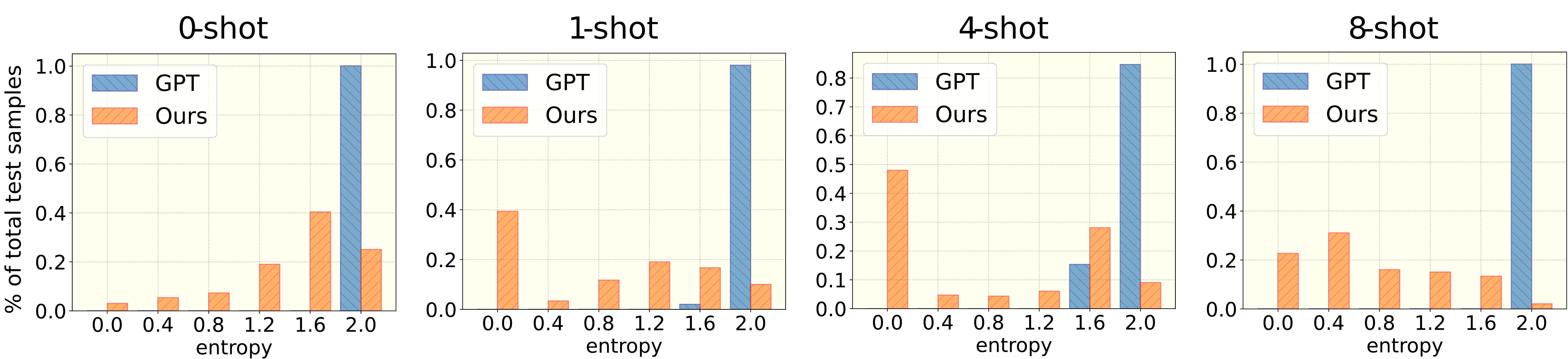}
    \vspace{-0.1cm}
    \caption{\small Shannon entropy histograms of using vanilla ICL on GPT-2-XL (1.5B) vs our method on AGNews (higher entropy implies higher uncertainty); we use logarithmic base two. Refer to Section~\ref{sec:shannon} for a detailed explanation.}
    \vspace{-0.1cm}
\end{figure*}

\begin{figure*}[h!]
    \centering
    \includegraphics[width=1.0\textwidth]{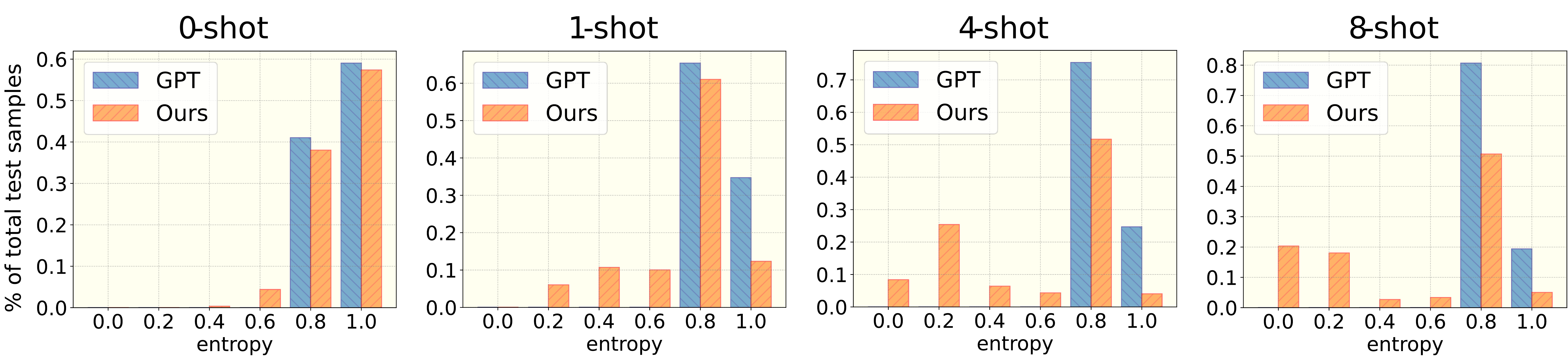}
    \vspace{-0.1cm}
    \caption{\small Shannon entropy histograms of using vanilla ICL on Llama-2 (13B) vs our method on SST-2 (higher entropy implies higher uncertainty); we use logarithmic base two. Refer to Section~\ref{sec:shannon} for a detailed explanation.}
    \vspace{-0.1cm}
\end{figure*}

\begin{figure*}[h!]
    \centering
    \includegraphics[width=1.0\textwidth]{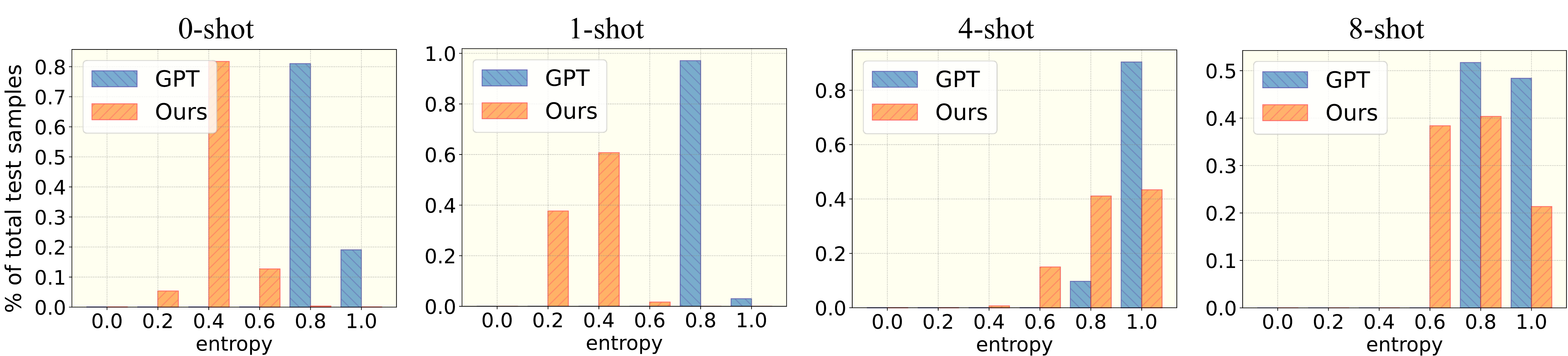}
    \vspace{-0.1cm}
    \caption{\small Shannon entropy histograms of using vanilla ICL on Llama-2 (13B) vs our method on Subj (higher entropy implies higher uncertainty); we use logarithmic base two. Refer to Section~\ref{sec:shannon} for a detailed explanation.}
    \vspace{-0.1cm}
\end{figure*}

\begin{figure*}[h!]
    \centering
    \includegraphics[width=1.0\textwidth]{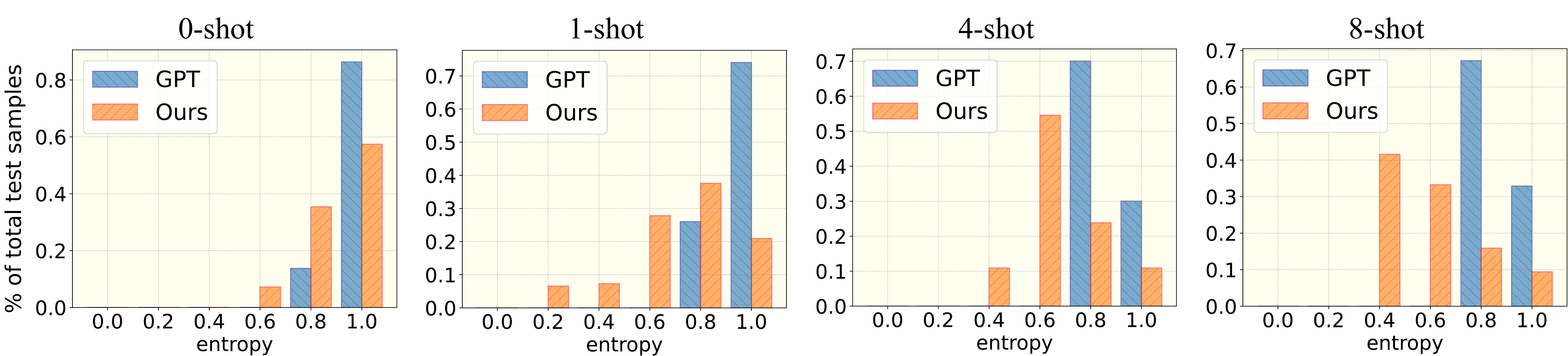}
    \vspace{-0.1cm}
    \caption{\small Shannon entropy histograms of using vanilla ICL on Llama-2 (13B) vs our method on RTE (higher entropy implies higher uncertainty); we use logarithmic base two. Refer to Section~\ref{sec:shannon} for a detailed explanation.}
    \vspace{-0.1cm}
\end{figure*}

\begin{figure*}[h!]
    \centering
    \includegraphics[width=1.0\textwidth]{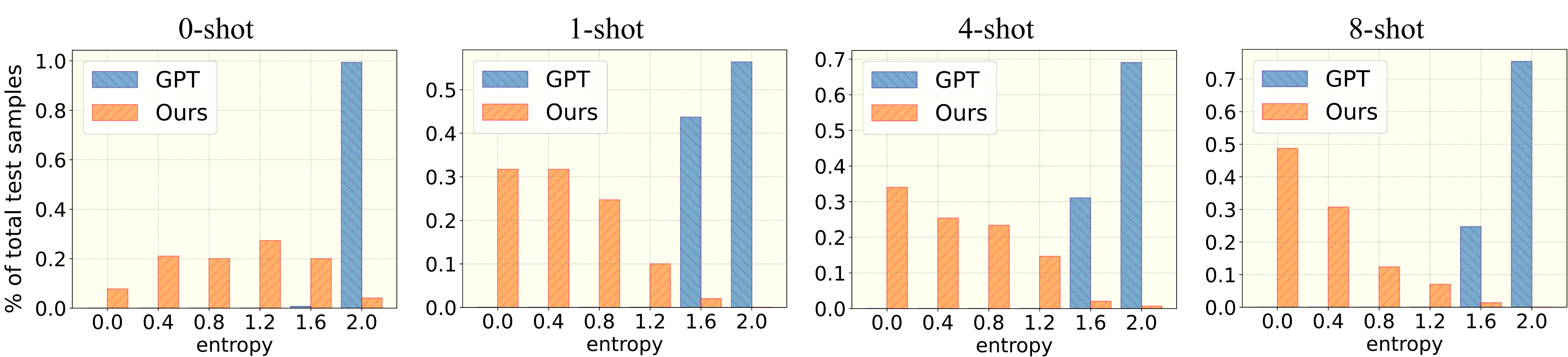}
    \vspace{-0.1cm}
    \caption{\small Shannon entropy histograms of using vanilla ICL on Llama-2 (13B) vs our method on AGNews (higher entropy implies higher uncertainty); we use logarithmic base two. Refer to Section~\ref{sec:shannon} for a detailed explanation.}
    \vspace{-0.1cm}
\end{figure*}

\begin{figure}[h]
\vspace{-0.1cm}
\begin{center}
\includegraphics[width=0.45\textwidth]{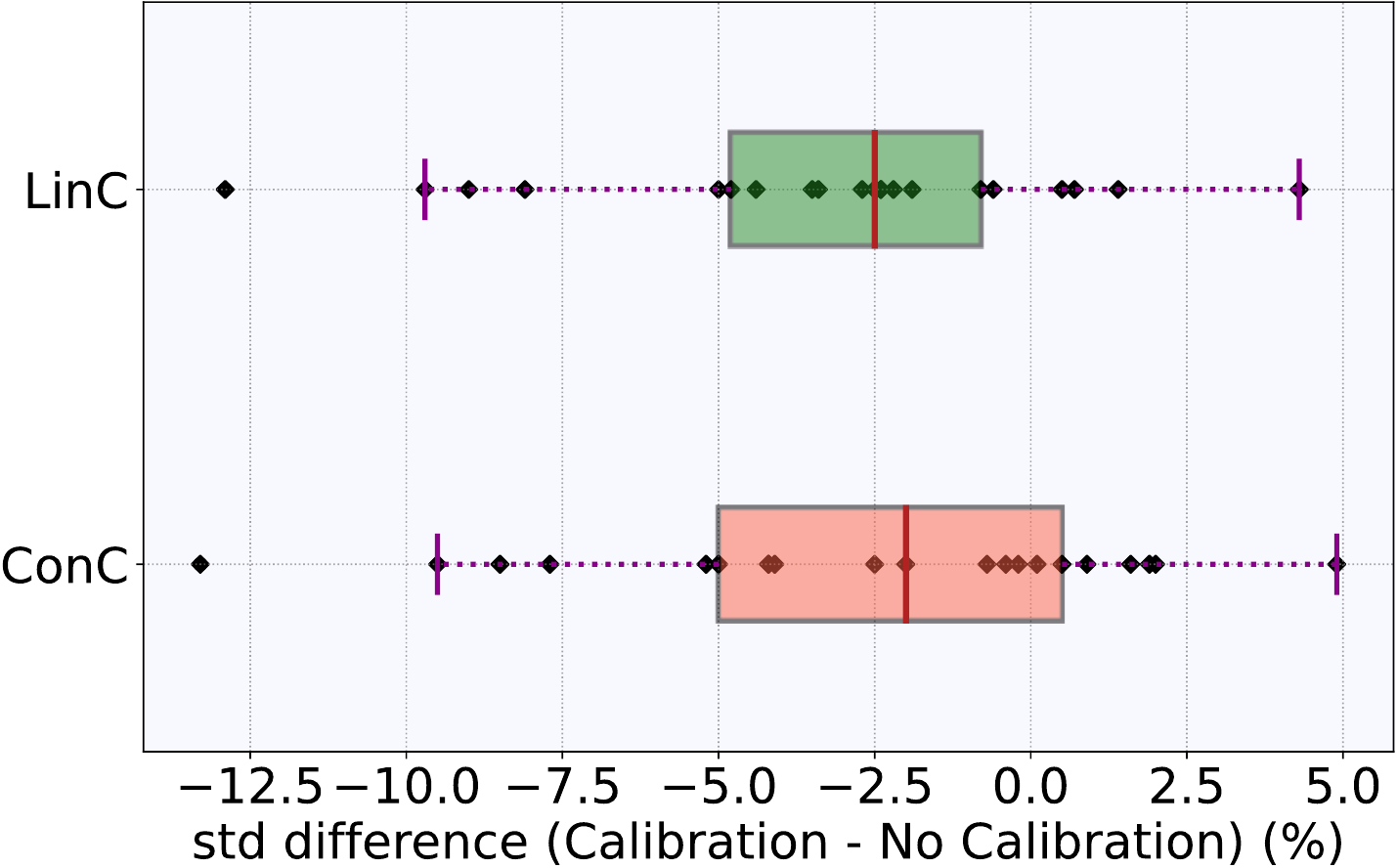}
  \end{center}
\vspace{-0.1cm}
\caption{\small LinC diminishes the standard deviation of accuracy across different demonstrations. The difference in standard deviation between the calibration methods and the unadjusted baseline (NoC) from Table \ref{table_GPTJ} is plotted.}
\label{box_plot_std}
\vspace{-0.2cm}
\end{figure}

\begin{table*}[tb]
  \renewcommand{\arraystretch}{1.0}
  \setlength\tabcolsep{2.7pt}
  \centering
  \begin{tabular}{ccllllllllll}
    \toprule
      Shot & Method & SST-2     & SST-5   & AGNews  &TREC &DBpedia & RTE  & Subj   & Avg   \\
    \midrule
    \specialrule{0em}{1pt}{1pt}
    \multicolumn{9}{c}{\emph{\footnotesize{GPT-J 6B}}}\\
    \specialrule{0em}{0pt}{1pt}
    \midrule

    \multirow{3}{*}{0-shot} & No Calibration &  66.3$_{0.0}$ & 33.7$_{0.0}$ & 36.0$_{0.0}$ & 24.7$_{0.0}$ & 19.7$_{0.0}$ & 55.6$_{0.0}$ & {65.7$_{0.0}$} & {43.1} \\
     & Contextual Calibration & {58.0$_{0.0}$} & 40.7$_{0.0}$ &  {56.0}$_{0.0}$ & 40.0$_{0.0}$ & 48.0$_{0.0}$ & 52.4$_{0.0}$ & {58.7$_{0.0}$} & {50.5}\\
     & Prototypical Calibration$^\ddagger$ & {74.2$_{0.2}$} & {42.1$_{0.8}$} &  {55.1}$_{0.4}$ & 53.4$_{6.1}$ & 66.1$_{1.5}$ & \textbf{57.0$_{1.0}$} & {69.5$_{0.2}$} & {59.6}\\
     & Prototypical Calibration$^*$ & {74.1$_{1.1}$} & {32.2$_{3.8}$} &  {49.8}$_{4.3}$ & 44.7$_{6.7}$ & 20.7$_{4.7}$ & {56.5$_{1.5}$} & {66.8$_{1.8}$} & {49.3}\\
     & \ours{} &  \textbf{{74.3}$_{0.0}$} &  \textbf{{46.0}$_{0.0}$} & \textbf{64.3$_{0.0}$} &  \textbf{{70.7}$_{0.0}$} &  \textbf{{69.3}$_{0.0}$} &  {{56.7}$_{0.0}$} & \textbf{70.7$_{0.0}$} &  \textbf{64.6} \\
     
     \specialrule{0em}{2pt}{1pt}
     
     \multirow{3}{*}{1-shot} & No Calibration & 67.3$_{6.7}$ & 35.3$_{3.5}$ & 65.3$_{14.3}$ & 40.3$_{8.8}$  & 64.9$_{16.6}$ & 50.7$_{3.7}$ & 65.1$_{9.9}$ & {55.6}\\
     & Contextual Calibration & 88.3$_{1.7}$ & 46.9$_{3.1}$ & 74.9$_{5.8}$ & 62.6$_{4.7}$ & 80.2$_{3.3}$ &  {53.4}$_{1.2}$ & 59.1$_{2.2}$ & {66.5}\\
     & Prototypical Calibration$^\ddagger$ & \textbf{90.8$_{1.7}$} & {47.6$_{2.5}$} &  {79.8}$_{5.4}$ & 55.3$_{6.4}$ & \textbf{90.0$_{2.2}$} & {56.7$_{3.1}$} & \textbf{77.9$_{4.8}$} & \textbf{71.2}\\
     & Prototypical Calibration$^*$ & {89.3$_{1.9}$} & {35.2$_{8.9}$} &  {78.1}$_{7.5}$ & 45.9$_{9.6}$ & {62.2$_{4.9}$} & \textbf{57.7$_{1.8}$} & {74.9$_{5.9}$} & {63.2}\\
     & \ours{} &  {{88.5}$_{1.7}$} &  \textbf{{50.4}$_{1.3}$} &  \textbf{{81.7}$_{4.6}$} &  \textbf{{63.0}$_{4.4}$} &  {{82.6}$_{3.7}$} & {56.0$_{1.8}$} & {69.9$_{7.5}$} &  {70.3}\\
     
     \specialrule{0em}{2pt}{1pt}

     \multirow{3}{*}{4-shot} & No Calibration & 88.9$_{3.3}$ & 46.3$_{3.2}$ & 72.3$_{6.1}$ & 37.7$_{4.2}$ & 82.1$_{14.4}$ & 55.0$_{7.1}$ & 57.4$_{6.9}$ & {62.8}\\
     & Contextual Calibration & 92.8$_{3.1}$ & 49.4$_{4.8}$ & 75.2$_{4.1}$ & 46.0$_{5.1}$ & 88.9$_{4.9}$ & 56.0$_{1.9}$ & 65.3$_{11.8}$ & {67.7}\\
     & Prototypical Calibration$^\ddagger$ & \textbf{95.0$_{0.4}$} & {46.2$_{4.6}$} &  {79.9}$_{6.6}$ & 57.1$_{5.3}$ & \textbf{91.9$_{2.6}$} & \textbf{61.2$_{2.7}$} & \textbf{79.4$_{5.8}$} & \textbf{73.0}\\
     & Prototypical Calibration$^*$ & {94.6$_{0.6}$} & {47.9$_{5.5}$} &  \textbf{81.5}$_{2.0}$ & 49.0$_{10.2}$ & {63.8$_{4.3}$} & {60.8$_{3.1}$} & {78.9$_{8.0}$} & {68.1}\\
     & \ours{} &  {{94.9}$_{0.9}$} &  \textbf{{51.1}$_{3.7}$} &  {{77.9}$_{5.5}$} &  \textbf{{65.3}$_{1.5}$} &  {{89.4}$_{5.4}$} &  {{58.4}$_{3.7}$} & {68.4$_{11.2}$} &  {72.2}\\
     
    \specialrule{0em}{2pt}{1pt}
    
    \multirow{3}{*}{8-shot} & No Calibration & 91.8$_{6.0}$ & 44.5$_{4.3}$ & 76.8$_{9.9}$ & 43.5$_{6.3}$ & 88.6$_{3.2}$ & 60.3$_{2.8}$ & {82.2$_{5.7}$} & {69.7}\\
     & Contextual Calibration & 93.8$_{1.8}$ &  {44.4}$_{4.4}$ & 78.5$_{5.7}$ & 52.3$_{8.3}$ & {90.8$_{2.5}$} & 58.8$_{4.7}$ & 81.3$_{6.2}$ & {71.4}\\
     & Prototypical Calibration$^\ddagger$ & {94.4$_{1.0}$} & {47.4$_{4.4}$} &  \textbf{84.2}$_{1.8}$ & 61.0$_{7.6}$ & \textbf{95.1$_{0.5}$} & {61.7$_{7.2}$} & {83.6$_{4.2}$} & {75.3}\\
     & Prototypical Calibration$^*$ & {94.1$_{1.9}$} & {46.8$_{8.7}$} &  {83.7}$_{1.4}$ & 47.5$_{8.3}$ & {68.5$_{6.8}$} & {61.4$_{4.7}$} & {83.8$_{5.9}$} & {69.4}\\
     & \ours{}  & \textbf{{94.8}$_{1.2}$} & \textbf{51.3$_{0.8}$} &  {{83.7}$_{1.8}$} &  \textbf{{67.3}$_{3.8}$} &  {{90.1}$_{3.9}$} &  \textbf{{63.2}$_{4.2}$} & \textbf{84.7$_{4.9}$}  &  \textbf{76.4} \\
     
    \bottomrule
  \end{tabular}
  
  \setlength{\abovecaptionskip}{0.2cm} 
  \caption{\small Performance comparisons among the conventional approach (No Calibration; \cite{GPT3}), Contextual Calibration \cite{zhao2021calibrate}, Prototypical Calibration \cite{han2022prototypical} and Linear Probe Calibration (\textit{Ours}) on GPT-J 6B. $^\ddagger$ denotes reporting the numbers from the original paper that utilize orders of magnitude larger number of samples (see Table \ref{table_samples_Proca}); $^*$ denotes our reproduced results that use same number of samples as used by our method. We report the mean and the standard deviation of test accuracy across different choices of the demonstrations (the prompt template is fixed). Prototypical Calibration's 0-shot accuracy deviation results from varying estimate sets across five random seeds \cite{han2022prototypical}. We also report the average performance across seven datasets.}
  \label{table_GPTJ-Proca}
\end{table*}

\begin{table*}[tb]
  \renewcommand{\arraystretch}{1.0}
  \setlength\tabcolsep{2.7pt}
  \centering
  \begin{tabular}{ccllllllllll}
    \toprule
      Shot & Method & SST-2     & SST-5   & AGNews  &TREC &DBpedia & RTE  & Subj  \\
    \midrule
    \specialrule{0em}{1pt}{1pt}
    \multicolumn{9}{c}{\emph{\footnotesize{GPT-J 6B}}}\\
    \specialrule{0em}{0pt}{1pt}
    \midrule

    \multirow{3}{*}{0-shot}
     & Prototypical Calibration & 500 & 2000 & 2000 & 2000 & 3000 & 1000 & 1000 \\
     & \ours{} &  100 &  30 & 300 &  300 &  300 &  100 & 100 \\
     
     \specialrule{0em}{2pt}{1pt}
     
     \multirow{3}{*}{1-shot} 
     & Prototypical Calibration & 500 & 2000 & 2000 & 2000 & 3000 & 1000 & 1000 \\
     & \ours{} &  300 &  30 & 300 &  100 &  30 &  100 & 100 \\ 
     
     \specialrule{0em}{2pt}{1pt}

     \multirow{3}{*}{4-shot} 
     & Prototypical Calibration & 500 & 2000 & 2000 & 2000 & 3000 & 1000 & 1000 \\
     & \ours{} &  100 &  100 & 100 &  100 &  30 &  100 & 100 \\ 
     
    \specialrule{0em}{2pt}{1pt}
    
    \multirow{3}{*}{8-shot}
     & Prototypical Calibration & 500 & 2000 & 2000 & 2000 & 3000 & 1000 & 1000 \\
     & \ours{} &  100 &  100 & 100 &  100 &  30 &  100 & 100 \\ 
     
    \bottomrule
  \end{tabular}
  
  \setlength{\abovecaptionskip}{0.2cm} 
  \caption{\small Validation sample size $N_{\rm v}$ used for different datasets}
  \label{table_samples_Proca}
\end{table*}

\begin{table*}[tb]
  \renewcommand{\arraystretch}{1.0}
  \setlength\tabcolsep{2.7pt}
  \centering
  \begin{tabular}{ccllllllllll}
    \toprule
      Method & SST-2     & SST-5   & AGNews  & RTE   \\
     
    \midrule

    \specialrule{0em}{1pt}{1pt}
    \multicolumn{5}{c}{\emph{\footnotesize{GPT-J 6B}}}\\
    \specialrule{0em}{0pt}{1pt}
    \midrule
     Zero Initialization &  74.7$_{0.0}$ & 43.0$_{0.0}$ & 65.0$_{0.0}$ & {56.7$_{0.0}$} \\
     Random Initialization & {74.7$_{0.0}$} & 43.0$_{0.0}$ &  {65.0}$_{0.0}$ & {56.7$_{0.0}$} \\
     Initialize via Eq. (5) &  {{74.3}$_{0.0}$} &  {{46.0}$_{0.0}$} & {64.3$_{0.0}$} & {56.7$_{0.0}$}  \\

    \bottomrule
  \end{tabular}
  
  \setlength{\abovecaptionskip}{0.2cm} 
  \caption{\small Performance comparisons among different initialization methods on 0-shot setting.}
  \label{table_init_comp}
\end{table*}

\textbf{Hyperparameter search.} After finding $N_v$ and $T$ via random search using the sets defined in Section \ref{section_4.2}, we narrowly fine-tune the learning rate $\alpha$ for each experiment from the range $[1e-5,2e1]$. Therefore, in each experiment, the value of $\alpha$ might vary, and to ensure transparency and reproducibility, we have compiled some hyperparameter sets in the $\rm {examples\_ssh}$ folder within the provided code. 

Furthermore, given that the calibration network constitutes a basic linear convex problem, determining an appropriate learning rate presents no issue, as the fixed rate can readily be substituted with a schedule that gradually reduces the learning rate as training advances, based on specified criteria (e.g., a linear scheduler that gradually diminishes the rate for each parameter group by applying a small multiplicative factor until a predetermined epoch count is attained). 

\textbf{Where is the affine transformation applied?} The classification tasks we considered apply the affine transformation to the set of probabilities that are associated with each class in the label space i.e. task-specific tokens. In other words, after we get the logits which are in the dimension (batch\_size, seq\_length, vocab\_size), we sample the tokens that correspond to the classes in the label space to a get a probability vector of dimension (batch\_size, num\_classes). We then apply the affine transformation on this probability vector.

\begin{wraptable}{r}{7.6cm}
\footnotesize
  \renewcommand{\arraystretch}{1.0}
  \setlength\tabcolsep{2.7pt}
  \centering
  \begin{tabular}{cclllllllllll}
    \toprule
      Model & Shots & Method & SST-5   & AGNews  \\

    \midrule
    \specialrule{0em}{1pt}{1pt}
    \specialrule{0em}{0pt}{1pt}

    \multirow{12}{*}{Llama-2 7B} 
    & \multirow{3}{*}{0-shot} &  NoC &  32.3$_{0.0}$ & 57.3$_{0.0}$  \\
    & & ConC & 35.0$_{0.0}$ & 63.3$_{0.0}$  \\
    & & LinC & \textbf{45.0$_{0.0}$} & \textbf{82.0$_{0.0}$}    \\
     
     \specialrule{0em}{2pt}{1pt}
     
     & \multirow{3}{*}{1-shot} & NoC & 39.3$_{9.8}$ & 83.9$_{2.1}$  \\
     & & ConC &  43.0$_{1.1}$ & 82.0$_{3.3}$   \\
     & & LinC & \textbf{50.1$_{3.5}$} & \textbf{84.3$_{2.6}$}  \\
     
     \specialrule{0em}{2pt}{1pt}
     
     & \multirow{3}{*}{4-shot} & NoC & 49.7$_{1.9}$ & 87.2$_{1.1}$   \\
     & & ConC & 44.3$_{2.5}$ & 87.2$_{1.2}$ \\
     & & LinC &  \textbf{53.5$_{2.8}$} & \textbf{87.6$_{1.1}$} \\
     
    \specialrule{0em}{2pt}{1pt}
    
     & \multirow{3}{*}{8-shot} & NoC & 48.9$_{5.5}$ & 86.7$_{0.9}$  \\
     & & ConC & 48.7$_{2.2}$ & 87.5$_{0.5}$  \\
     & & LinC  & \textbf{52.0$_{3.3}$} & \textbf{87.6$_{0.6}$}  \\

     \midrule
     \multirow{12}{*}{Llama-2 13B} 
    & \multirow{3}{*}{0-shot} &  NoC &  34.0$_{0.0}$ & 73.3$_{0.0}$  \\
    & & ConC & 33.0$_{0.0}$ & 72.3$_{0.0}$  \\
    & & LinC & \textbf{47.3$_{0.0}$} & \textbf{85.7$_{0.0}$}    \\
     
     \specialrule{0em}{2pt}{1pt}
     
     & \multirow{3}{*}{1-shot} & NoC & 43.9$_{3.5}$ & 81.5$_{2.7}$  \\
     & & ConC &  42.4$_{3.9}$ & 80.6$_{1.9}$   \\
     & & LinC & \textbf{51.0$_{1.6}$} & \textbf{84.1$_{1.7}$}  \\
     
     \specialrule{0em}{2pt}{1pt}
     
     & \multirow{3}{*}{4-shot} & NoC & 48.7$_{4.1}$ & 82.7$_{3.7}$   \\
     & & ConC & 44.9$_{5.2}$ & 80.5$_{2.3}$ \\
     & & LinC &  \textbf{53.0$_{0.9}$} & \textbf{86.0$_{1.2}$} \\
     
    \specialrule{0em}{2pt}{1pt}
    
     & \multirow{3}{*}{8-shot} & NoC & 49.1$_{7.3}$ & 87.0$_{0.5}$  \\
     & & ConC & 47.1$_{4.4}$ & 83.8$_{1.6}$  \\
     & & LinC  & \textbf{51.4$_{6.1}$} & \textbf{87.1$_{0.6}$}  \\

    \bottomrule
  \end{tabular}
  
  \setlength{\abovecaptionskip}{0.2cm} 
  \vspace{-0.1cm}
  \caption{\small Performance under same model family (i.e. Llama-2) but different sizes (7B/13B).}
  \label{table_Llama7B}
        \vspace{-2cm}
\end{wraptable}
\textbf{Expected Calibration Error (ECE) \cite{Naeini2015}.} The ECE computes the Expected Calibration Error across the bins in the following manner:
\begin{align}
    ECE = \sum_{m=1}^{M} \frac{|B_{m}|}{n} |o_m - e_m|
\end{align}
where $o_m$ represents the accuracy i.e. the true fraction of positive instances in bin $m$, $e_m$ represents the model confidence i.e. 
is the mean of the post-calibrated probabilities for the instances in bin $m$, and the fraction $\frac{B_m}{n}$ denotes the empirical probability (fraction) of all instances that fall into bin. Therefore, The ECE can be seen as evaluating the extent to which a model's estimated "probabilities" align with the actual (observed) probabilities by computing a weighted average of the absolute difference between accuracy and confidence.  A model's calibration is considered better when ECE values are lower and an ECE of zero is indicative of a perfectly calibrated model.

\subsection{Experiment Settings} \label{PEFT_settings}
For LoRA, we used a step size of 1.5e-6, a batch size of 16, a scaling factor of 32, and incorporated a 0.1 dropout probability within the LoRA layers. For SPT, we used a step size of 3.5e-4, a batch size of 16, specified 8 virtual tokens (i.e. the num\_vir\_tokens argument), and employed an initial text for prompt tuning (i.e. the prompt\_tuning\_init\_text argument) that consisted of a list of classes separated by commas (for example, "World, Sports, Business, Technology" for AGNews). Across all our experiments, we performed fine-tuning over a total of 20 epochs.

For the non-language tasks that used tabular datasets from OpenML \cite{vanschoren2014openml}, following \cite{dinh2022lift}, we selected the maximum number of instances that could be accommodated within the contextual window for each dataset. Additionally, our results were reported as an average across three random seeds.

\begin{table*}[tb]
  \renewcommand{\arraystretch}{1.0}
  \setlength\tabcolsep{2.7pt}
  \centering
  \begin{tabular}{ccllllllllll}
    \toprule
     & Method  & 120  &  600 & 1.2k    & 6k  \\
    \midrule
    \specialrule{0em}{1pt}{1pt}
    \multicolumn{6}{c}{\emph{\footnotesize{GPT-J 6B}}}\\
    \specialrule{0em}{0pt}{1pt}
    \midrule

    \multirow{3}{*}{Ep$\#$1}
     & SPT & 27.7 & 26.3 & 43.3 & 78.7 \\
     & SPT+LinC &  66.3 &  59.7 & 73.3 &  83.0 \\
     & Difference &  +38.7 &  +33.3 & +30.0 &  +4.3 \\
     
     \specialrule{0em}{2pt}{1pt}
     
     \multirow{3}{*}{Ep$\#$3} 
     & SPT & 41.0 & 57.7 & 61.7 & 86.0 \\
     & SPT+LinC &  69.3 &  74.7 & 79.0 &  86.7  \\ 
     & Difference &  +28.3 &  +17.0 & +17.3 &  +0.7 \\
     
     \specialrule{0em}{2pt}{1pt}

     \multirow{3}{*}{Ep$\#$10} 
     & SPT        &  30.7  & 51.3 & 55.7 & 84.3 \\
     & SPT+LinC   &  42.3 &  74.7 & 77.7 &  86.0  \\ 
     & Difference &  +11.7 &  +23.3 & +22.0 &  +1.7 \\
     
    \specialrule{0em}{2pt}{1pt}
    
    \multirow{3}{*}{Ep$\#$20}
     & SPT & 29.7 & 58.7 & 65.3 & 84.0 \\
     & SPT+LinC &  34.7 &  77.0 & 77.7 &  86.3  \\ 
     & Difference &  +5.0 &  +18.3 & +12.3 &  +2.3 \\

     \specialrule{0em}{2pt}{1pt}
    
    \multirow{3}{*}{Best}
     & SPT & 41.0 & 62.3 & 71.0 & 86.0  \\
     & SPT+LinC &  69.3 &  80.3 & 81.3 &  86.7 \\ 
     & Difference &  +28.3 &  +18.0 & +10.3 &  +0.7 \\

    \midrule
    \specialrule{0em}{1pt}{1pt}
    \multicolumn{6}{c}{\emph{\footnotesize{Llama-2 13B}}}\\
    \specialrule{0em}{0pt}{1pt}
    \midrule

    \multirow{3}{*}{Ep$\#$1}
     & SPT & 43.0 & 44.7 & 55.0 & 63.3 \\
     & SPT+LinC &  80.3 &  84.7 & 85.0 &  84.0 \\
     & Difference &  +37.3 &  +40.0 & +30.0 &  +20.7 \\
     
     \specialrule{0em}{2pt}{1pt}
     
     \multirow{3}{*}{Ep$\#$3} 
     & SPT & 41.3 & 61.3 & 64.7 & 71.0 \\
     & SPT+LinC &  79.7 &  85.3 & 78.7 &  84.0  \\ 
     & Difference &  +38.4 &  +24.0 & +14.0 &  +13.0 \\
     
     \specialrule{0em}{2pt}{1pt}

     \multirow{3}{*}{Ep$\#$10} 
     & SPT        &  49.0  & 39.3 & 79.3 & 86.0 \\
     & SPT+LinC   &  84.3 &  69.3 & 84.0 &  86.3  \\ 
     & Difference &  +35.3 &  +30.0 & +4.7 &  +0.3 \\
     
    \specialrule{0em}{2pt}{1pt}
    
    \multirow{3}{*}{Ep$\#$20}
     & SPT & 62.0 & 79.0 & 84.0 & 84.7  \\
     & SPT+LinC &  79.3 &  84.7 & 84.7 &  85.3  \\ 
     & Difference &  +17.3 &  +5.7 & +0.7 &  +0.6 \\

     \specialrule{0em}{2pt}{1pt}
    
    \multirow{3}{*}{Best}
     & SPT & 62.0 & 79.0 & 84.0 & 87.3  \\
     & SPT+LinC &  79.3 &  84.7 & 84.7 &  87.3  \\ 
     & Difference &  +17.3 &  +5.7 & +0.7 &  0.0 \\
     
    \bottomrule
  \end{tabular}
  
  \setlength{\abovecaptionskip}{0.2cm} 
  \caption{\small We assess the impact of Soft Prompt Tuning (SPT) on GPT-J 6B and Llama-2 13B using AGNews dataset with and without LinC in two distinct scenarios: 1) high compute and abundant data (lower-right), and 2) limited compute and sparse data (upper-left). The columns represent the number of data points employed for SPT, while the rows represent the number of training epochs.}
  \label{table_SPT}
\end{table*}

\begin{table*}[tb]
  \renewcommand{\arraystretch}{1.0}
  \setlength\tabcolsep{2.7pt}
  \centering
  \begin{tabular}{ccllllllllll}
    \toprule
     & Method  & 120  &  600 & 1.2k    & 6k  \\
    \midrule
    \specialrule{0em}{1pt}{1pt}
    \multicolumn{6}{c}{\emph{\footnotesize{GPT-J 6B}}}\\
    \specialrule{0em}{0pt}{1pt}
    \midrule

    \multirow{3}{*}{Ep$\#$1}
     & LoRA       &  36.7  &  26.0  & 72.3   &  86.3 \\
     & LoRA+LinC  &  66.0  &  70.7  & 85.0   &  87.0 \\
     & Difference &  +29.3 &  +44.7 & +12.7  &  +0.7 \\
     
     \specialrule{0em}{2pt}{1pt}
     
     \multirow{3}{*}{Ep$\#$3} 
     & LoRA       &  26.3  &  67.0  & 80.0  &  83.0 \\
     & LoRA+LinC  &  66.3  &  85.0  & 84.7  &  88.0  \\ 
     & Difference &  +40.0 &  +18.0 & +4.7  &  +5.0 \\
     
     \specialrule{0em}{2pt}{1pt}

     \multirow{3}{*}{Ep$\#$10} 
     & LoRA        &  41.7 &  76.3  & 78.3  &  85.7 \\
     & LoRA+LinC   &  66.0 &  84.3  & 84.0  &  87.3  \\ 
     & Difference &  +24.3 &  +8.0 & +6.0   &  +1.6 \\
     
    \specialrule{0em}{2pt}{1pt}
    
    \multirow{3}{*}{Ep$\#$20}
     & LoRA       &  43.3  &  75.0  & 78.3  &  87.7 \\
     & LoRA+LinC  &  69.0  &  83.7  & 82.7  &  88.0  \\ 
     & Difference &  +25.7 &  +8.7 & +4.4   &  +0.3 \\

     \specialrule{0em}{2pt}{1pt}
    
    \multirow{3}{*}{Best}
     & LoRA       &  41.7  &  78.3  & 82.0  &  88.7  \\
     & LoRA+LinC  &  66.0  &  83.7  & 83.7  &  88.7 \\ 
     & Difference &  +24.3 &  +5.4 & +1.7   &  0.0 \\
     
    \midrule
    \specialrule{0em}{1pt}{1pt}
    \multicolumn{6}{c}{\emph{\footnotesize{Llama-2 13B}}}\\
    \specialrule{0em}{0pt}{1pt}
    \midrule

     \multirow{3}{*}{Ep$\#$1}
     & LoRA       &  73.3  &  80.0  & 86.7   &  83.3 \\
     & LoRA+LinC  &  85.7  &  86.0  & 87.3   &  87.0 \\
     & Difference &  +12.4 &  +6.0 & +0.6  &  +3.7 \\
     
     \specialrule{0em}{2pt}{1pt}
     
     \multirow{3}{*}{Ep$\#$3} 
     & LoRA       &  73.7  &  84.0  & 86.3  &  85.7 \\
     & LoRA+LinC  &  85.7  &  87.3  & 87.7  &  88.0  \\ 
     & Difference &  +12.0 &  +3.3 & +1.4 &  +2.3 \\
     
     \specialrule{0em}{2pt}{1pt}

     \multirow{3}{*}{Ep$\#$10} 
     & LoRA        &  73.7 &  86.3  & 85.0  &  88.0 \\
     & LoRA+LinC   &  85.3 &  87.0  & 86.0  &  89.3  \\ 
     & Difference &  +11.6 &  +0.7 & +1.0   &  +1.3 \\
     
    \specialrule{0em}{2pt}{1pt}
    
    \multirow{3}{*}{Ep$\#$20}
     & LoRA       &  73.3  &  86.3  & 85.3  &  86.7 \\
     & LoRA+LinC  &  85.0  &  87.0  & 85.7  &  87.7  \\ 
     & Difference &  +11.7 &  +0.7 & +0.4   &  +1.0 \\

     \specialrule{0em}{2pt}{1pt}
    
    \multirow{3}{*}{Best}
     & LoRA       &  73.7  &  87.0  & 86.7  &  89.0 \\
     & LoRA+LinC  &  85.7  &  88.3  & 87.3  &  89.0 \\ 
     & Difference &  +12.0 &  +1.3 & +0.6   &  0.0 \\
     
    \bottomrule
  \end{tabular}
  
  \setlength{\abovecaptionskip}{0.2cm} 
  \caption{\small We assess the impact of Low-Rank Adaptation (LoRA) on GPT-J 6B and Llama-2 13B using AGNews dataset with and without LinC in two distinct scenarios: 1) high compute and abundant data (lower-right), and 2) limited compute and sparse data (upper-left). The columns represent the number of data points employed for LoRA, while the rows represent the number of training epochs.}
  \label{table_LoRA}
\end{table*}

\section{Difference from Contextual Calibration (ConC).}  \label{LinC_vs_ConC}
Both ConC and our method LinC employ an affine transformation, however, the key difference is that \textbf{LinC learns the transformation parameters using a few additional samples} (in the same format utilized by instruction-tuning). Conversely, ConC \textbf{does not learn} and instead employs a pre-defined initialization. 
More specifically, they create a prompt $P_{\rm cf} = P( \text{"NA"}, (x_i, y_i)_{i=1}^{k})$ and obtain $\mathbf{p}$ for this content-free input, denoted by $\mathbf{p}_{\sf cf}$ i.e. $\mathbf{p}_{\sf cf} = \mathbf{p}(P_{\rm cf})$. The parameters are set via: 
\begin{align} 
     \mathbf{A} = {\sf diag}(\mathbf{p}_{\sf cf})^{-1} ~~~{\rm and}~~~\mathbf{b}= \mathbf{0}.
     \label{Eq_cf}
\end{align}
They use Equation \eqref{Eq_cf} to \textbf{hard-code these parameters}.
On the other hand, LinC starts with a set of calibration parameters where zero initilization is one of the possible initialization method, see Table \ref{table_init_comp}. Then LinC utilizes a few additional data in the form of prompts to train the matrix $\mathbf{A}$ and vector $\mathbf{b}$ before applying the affine transformation. Note that LinC requires \textbf{orders of magnitue less data than the existing calibration methods} such as ProCa \cite{han2022prototypical}. For more in-depth comparison results, please refer to Table \ref{table_samples_Proca}.
Remarkably, in some tasks it suffices for LinC to use \textbf{as few as only 5 additional samples} (see Figure \ref{ablation_samples}).

\section{CO2 Emission Related to Experiments} \label{section_Co2_emissions}
Experiments in this paper were conducted using Amazon Web Services in region us-east-1, which has a carbon efficiency of 0.37 kgCO$_2$eq/kWh. For each experiment, the total emissions are estimated to be 0.055 kgCO$_2$eq, which are calculated using the \href{https://mlco2.github.io/impact#compute}{MachineLearning Impact calculator}   in \cite{lacoste2019quantifying}. 
While training or fine-tuning an LLM can result in significantly larger carbon emissions, with GPT-3 training requiring around 500,000 kgCO$_2$eq \cite{patterson2021carbon} and GPT-2-XL estimated to emit between 100,000 and 200,000 kgCO$_2$eq, our method trains low-dimensional parameters with minimal carbon emissions that are comparable to the inference stage. This makes our approach environmentally friendly in comparison to other methods.

\begin{table}[h]
  \begin{center}
  \small
  \begin{tabular}{|c|c|c|c|c|c|c|c|c|c|c|}
    \hline
    \multirow{5}{0.1cm} \textbf{Dataset}  & \multicolumn{3}{c|}{\textbf{Llama-2 (13B)}} \\
    \cline{2-4}
    &  \textbf{ICL} & \textbf{ConC} & \textbf{LinC}  \\
    \hline
    SST-2               & 0.1766 & \textbf{0.0943} & \underline{0.1263 } \\ \hline  \!\! \!\!   
    SST-5 \!\! \!\!     & 0.1760 & \underline{0.1681} & \textbf{0.1095}   \\ \hline
    AGNews              & 0.1294 & \textbf{0.0791} & \underline{0.0939}   \\ \hline
\!\! \!\!    
    TREC  \!\!  \!\!    & 0.2141 & \underline{0.1823} & \textbf{0.0782}   \\ \hline
    DBPedia             & 0.2000 & \underline{0.1744} & \textbf{0.1124}   \\ \hline
    RTE                 & \textbf{0.0369} & \underline{0.0467} & 0.0600    \\ \hline
    Subj                & 0.3183 & \underline{0.2658} & \textbf{0.1193}    \\ \hline
  \end{tabular}
  \caption{Expected Calibration Error (ECE) comparison between baselines and LinC on Llama-2 (13B) under 0-shot setting (a model with perfect calibration would exhibit an ECE of 0).}
  \label{table_ECE_llama2}
  \end{center}
  \vspace{-0.3cm}
\end{table}

\end{document}